\documentclass[sigconf]{acmart}

\usepackage{booktabs} 
\usepackage{subfigure}
\usepackage{url}
\usepackage{multirow}
\usepackage{multicol}
\usepackage{graphicx}
\usepackage{enumerate}
\usepackage{bm}

\newcommand{\tabincell}[2]{\begin{tabular}{@{}#1@{}}#2\end{tabular}}

\graphicspath{
    {./images/}{./images/proj/}{./images/weight/}{./images/scatter/}
}

\copyrightyear{2018}
\acmYear{2018}
\setcopyright{acmcopyright}
\acmConference[MM '18]{2018 ACM Multimedia Conference}{October 22--26,
2018}{Seoul, Republic of Korea}
\acmBooktitle{2018 ACM Multimedia Conference (MM '18), October 22--26,
2018, Seoul, Republic of Korea}
\acmPrice{15.00}
\acmDOI{10.1145/3240508.3240581}
\acmISBN{978-1-4503-5665-7/18/10}

\begin{document}
\title{Bridge the Gap Between VQA and Human Behavior on Omnidirectional Video}
\subtitle{A Large-Scale Dataset and a Deep Learning Model}

\author{Chen Li, Mai Xu*, Xinzhe Du, Zulin Wang}

\affiliation{%
  \institution{School of Electronic and Information Engineering, Beihang University (BUAA), Beijing, China}
  \country{*Corresponding author: MaiXu@buaa.edu.cn}
}

%
%
%

\begin{abstract}
Omnidirectional video enables spherical stimuli with the $360 \times 180^ \circ$ viewing range.
Meanwhile, only the viewport region of omnidirectional video can be seen by the observer through head movement (HM), and an even smaller region within the viewport can be clearly perceived through eye movement (EM).
Thus, the subjective quality of omnidirectional video may be correlated with HM and EM of human behavior.
To fill in the gap between subjective quality and human behavior, this paper proposes a large-scale visual quality assessment (VQA) dataset of omnidirectional video, called VQA-OV, which collects 60 reference sequences and 540 impaired sequences.
Our VQA-OV dataset provides not only the subjective quality scores of sequences but also the HM and EM data of subjects.
By mining our dataset, we find that the subjective quality of omnidirectional video is indeed related to HM and EM.
Hence, we develop a deep learning model, which embeds HM and EM, for objective VQA on omnidirectional video.
Experimental results show that our model significantly improves the state-of-the-art performance of VQA on omnidirectional video.
\end{abstract}

%
%



\maketitle
\begin{figure}[!tb]
\begin{center}
\includegraphics[width=.8\linewidth]{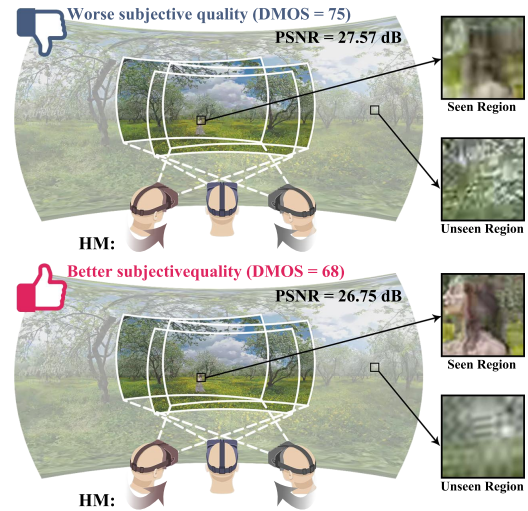}
\end{center}
\caption{An Example for the impaired omnidirectional sequences with similar VQA results in terms of peak signal-to-noise ratio (PSNR), but with different subjective quality in terms of differential mean opinion score (DMOS).}
\vspace{-1em}
\label{fig:1}
\end{figure}
\section{Introduction}\label{sec:intro}
Along with the rapid development of virtual reality (VR), omnidirectional video,  as a new type of multimedia, has been flooding into our daily life. Omnidirectional video enables spherical stimuli, which means that the whole $360\times180^{\circ}$ spherical space is accessible to human observers with the support of a head-mounted display (HMD).
The spherical stimuli of omnidirectional video bring immersive and interactive visual experience, but at the cost of extraordinarily high resolution. This tradeoff poses the technical challenges on storage, transmission, etc \cite{nasrabadi2017adaptive,corbillon2017optimal}.
Such challenges degrade the visual experience of omnidirectional video.
Therefore, it is necessary to study on visual quality assessment (VQA) for omnidirectional video.

Most recently, several subjective VQA approaches \cite{xu2017subjective, upenik2016testbed,zhang2017subjective,schatz2017towards,singla2017measuring,singla2017comparison,macquarrie2017cinematic,tran2017subjective} and objective VQA approaches \cite{xiu2017evaluation,tran2017evaluation,sun2017weighted,yu2015framework,zakharchenko2016quality,upenik2017performance} have been proposed for either omnidirectional image or video.
Among these approaches, the impact of some factors on the quality of omnidirectional video were studied, such as display types, coding schemes and sample uniformity under different map projections.
Figure 1 shows that the subjective visual quality of omnidirectional video is also related to the human behavior of head movement (HM). As shown in Figure 1, the impaired omnidirectional sequence with severer distortion (lower PSNR) is of better subjective quality (lower DMOS) because subjects rate scores on the basis of the seen region, which is determined by the HM of subjects.
However, none of the above approaches considers human behavior of viewing omnidirectional video, and thus they cannot well reflect the subjective quality of omnidirectional video.

When viewing omnidirectional video, humans are able to freely move their head to make their viewports focus on the attractive regions.
In other words, the regions outside the viewports cannot be observed by humans. Hence, HM is an important human behavior on viewing omnidirectional video, which significantly differs from traditional 2D video.
Additionally, eye movement (EM) decides which content within the viewport can be clearly captured at high resolution, similar to that of 2D video \cite{7742914}.
Hence, the human behavior of HM and EM is rather important in determining visual quality of omnidirectional video.
In fact, there are many latest works  \cite{corbillon2017360,wu2017dataset,rai2017dataset,rai2017saliency,de2017look,aladagli2017predicting,sitzmann2018saliency} concerning the human behavior of HM and EM in watching omnidirectional image/video.
Along with these approaches, some omnidirectional image/video datasets were established to collect HM or EM data.
However, to the best of our knowledge, there exists no work investigating how visual quality is related to human behavior on viewing omnidirectional video.
Even worse, there is no omnidirectional dataset that contains both subjective VQA scores and the corresponding HM/EM data, for studying the relationship between visual quality and human behavior.

To bridge the gap between VQA and human behavior on omnidirectional video, this paper establishes a large-scale VQA dataset of omnidirectional video (VQA-ODV), which is composed of subjective scores, HM data and EM data on 600 omnidirectional sequences.
It is worth mentioning that the 600 sequences in VQA-ODV are diverse in the content, duration and resolution, with impairments in both compression and map projection.
Then, we mine our dataset and find out that the subjective VQA scores of omnidirectional video are rather correlated with the HM and EM of observers.
Therefore, we propose a deep learning model, which incorporates both the HM and EM data of human behavior, for objective VQA on omnidirectional video.
The experimental results show that the HM and EM data embedded in our deep learning model are able to significantly advance the state-of-the-art performance of objective VQA on omnidirectional video.
This also verifies the effectiveness of considering human behavior in assessing visual quality of omnidirectional video.
\section{Related Works}\label{sec:works}
\textbf{VQA on omnidirectional video.}
In recent years, extensive works have emerged for VQA on omnidirectional video.
For subjective VQA, a handful of testbeds were proposed to subjectively rate the content \cite{upenik2016testbed} and streaming \cite{schatz2017towards} of omnidirectional video.
Additionally, some subjective experiments were conducted in \cite{macquarrie2017cinematic,tran2017subjective, singla2017measuring},
finding some key factors that have impact on the visual quality of omnidirectional video under different scenes \cite{macquarrie2017cinematic,tran2017subjective} and devices \cite{singla2017measuring}.
Additionally, several subjective VQA methods \cite{zhang2017subjective, singla2017comparison, xu2017subjective} were proposed.
For example, subjective assessment of multimedia panoramic video quality (SAMPVIQ) \cite{zhang2017subjective} and modified absolute category rating (M-ACR) \cite{singla2017comparison} were developed along with new subjective experiment procedures.
Moreover, a pair of overall DMOS (O-DMOS) and vectorized DMOS (V-DMOS) was proposed as a new subjective score processing method \cite{xu2017subjective}, in which the consistency of viewing direction across subjects is taken into account.

For objective VQA on omnidirectional video, there have been several approaches \cite{sun2017weighted, xiu2017evaluation, yu2015framework, zakharchenko2016quality, tran2017evaluation,upenik2017performance} that advance the metric of PSNR by considering the sample density of the map projection in omnidirectional video. Some applied weight allocation in the calculation of PSNR, e.g, weighted-to-spherically-uniform PSNR (WS-PSNR) \cite{sun2017weighted} and area weighted spherical PSNR (AW-SPSNR) \cite{xiu2017evaluation}; others resampled the content into a point-uniformed projection, e.g., S-PSNR \cite{yu2015framework} and PSNR in Craster parabolic projection (CPP-PSNR) \cite{zakharchenko2016quality}.
In \cite{tran2017evaluation,upenik2017performance}, the performance of the above objective VQA approaches was compared by measuring their correlation with the subjective quality scores.
Unfortunately, none of the above VQA approaches considers human behavior on viewing omnidirectional video, which significantly influences the quality of experience (QoE) \cite{Tao15, 7962230}.

\textbf{Human behavior on omnidirectional video.}
Dataset is fundamental in analyzing human behavior on viewing omnidirectional video.
Most recently, some omnidirectional image/video datasets have been established to collect the HM data \cite{corbillon2017360,wu2017dataset,xu2017subjective} and EM data \cite{rai2017dataset} of subjects.
Given these datasets, it is possible to analyze and model human behavior \cite{xu2017subjective, rai2017saliency} on viewing omnidirectional image/video.
Additionally, some works \cite{de2017look,aladagli2017predicting,sitzmann2018saliency} have been recently proposed to predict human's HM and EM on omnidirectional video, similar to saliency prediction on 2D video.
Although human behavior on omnidirectional video has been thoroughly studied, there exists no work investigating how visual quality is related to human viewing behavior. Meanwhile, there is no omnidirectional dataset that contains both subjective VQA scores and the corresponding HM/EM data in viewing omnidirectional video.
Thus, we establish the VQA-ODV dataset and then analyze on our dataset, to bridge the gap between VQA and human behavior on omnidirectional video.
We also implant HM and EM in a deep learning model to show that the performance of VQA can be significantly enhanced for omnidirectional video by considering human behavior.
\begin{figure*}[!tb]
\begin{center}
\includegraphics[width=1\textwidth]{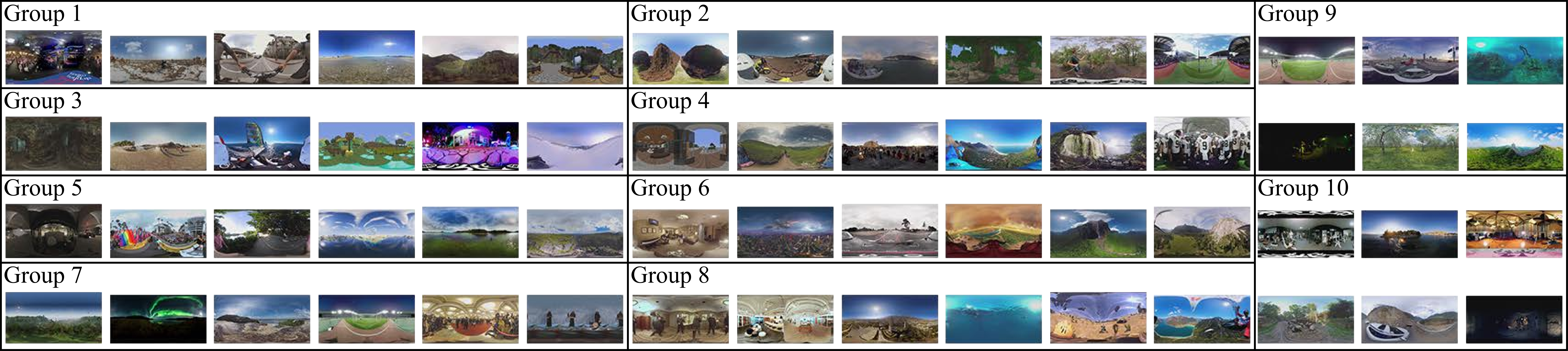}
\end{center}
\vspace{-1em}
\caption{All reference sequences in the VQA-ODV dataset. It can be seen that the categories of the content are diverse in each group.}
\label{fig:screen}
\end{figure*}
\begin{figure*}[!tb]
\begin{center}
\resizebox{\textwidth}{!}{
\hspace{-0.75em}
\subfigure[ERP]{
  \label{fig:proj:erp}
  \includegraphics[width=0.25\textwidth]{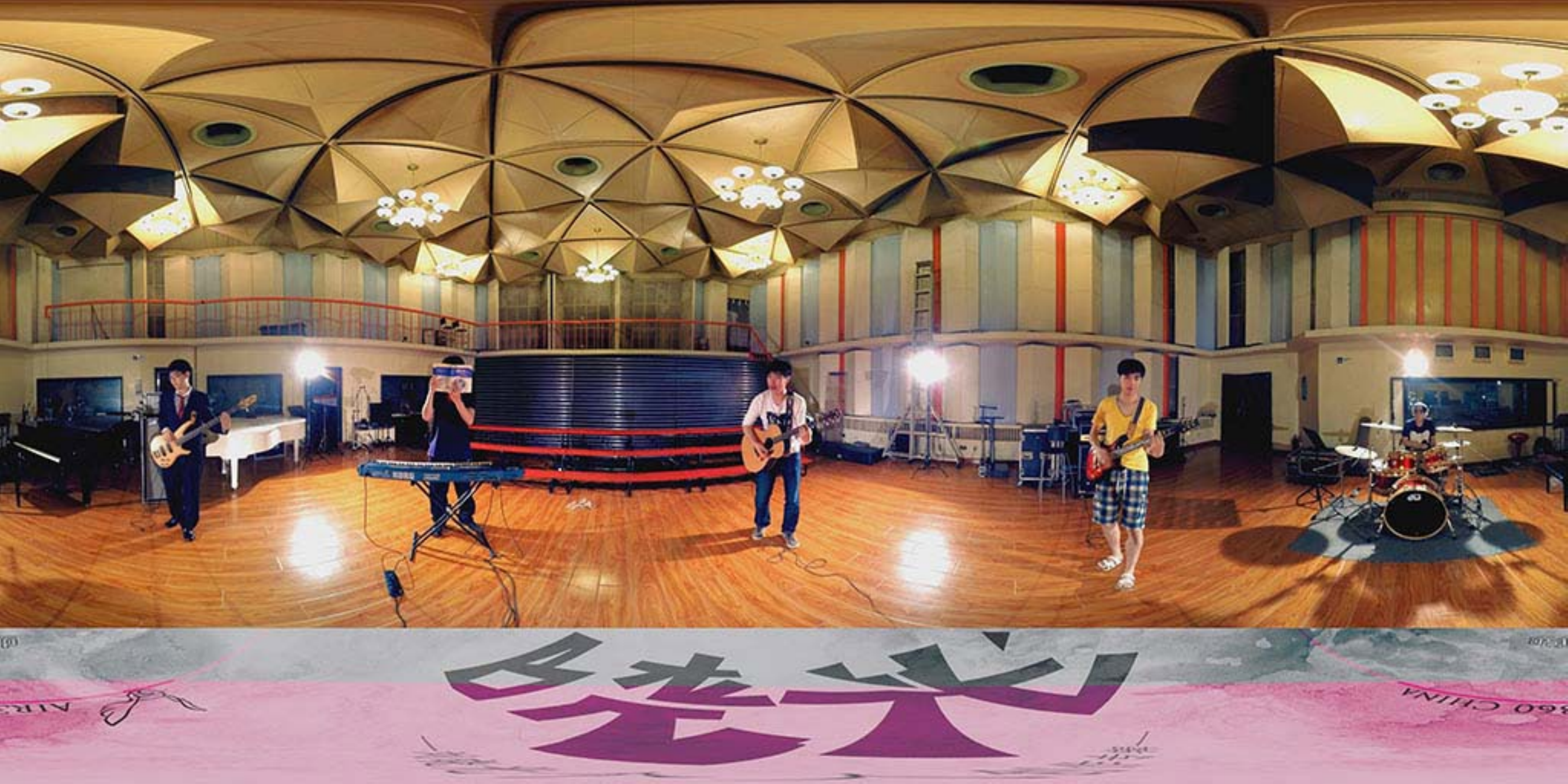}
}
\subfigure[RCMP]{
  \label{fig:proj:rcmp}
  \includegraphics[height=0.125\textwidth]{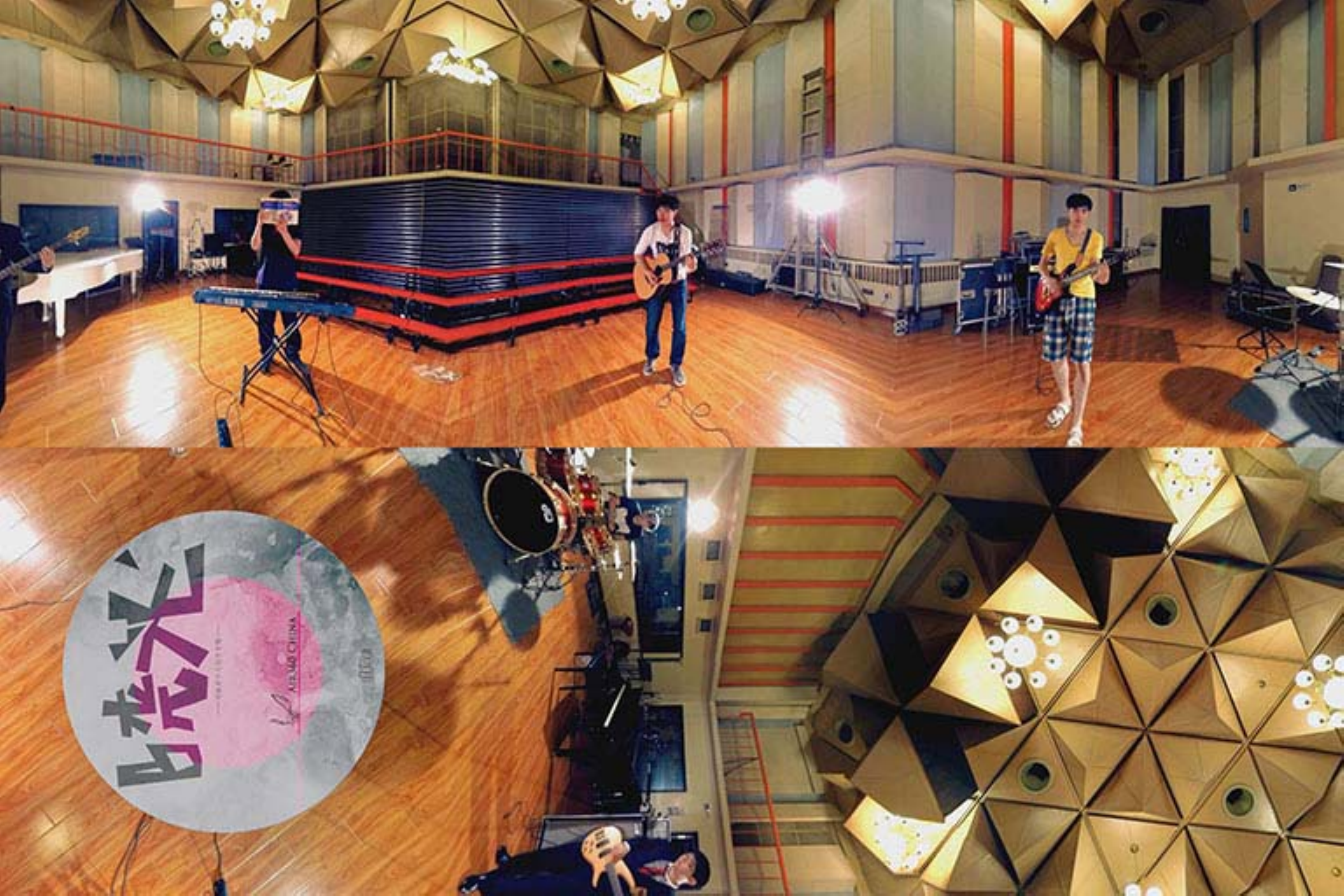}
}
\subfigure[TSP]{
  \label{fig:proj:tsp}
  \includegraphics[width=0.25\textwidth]{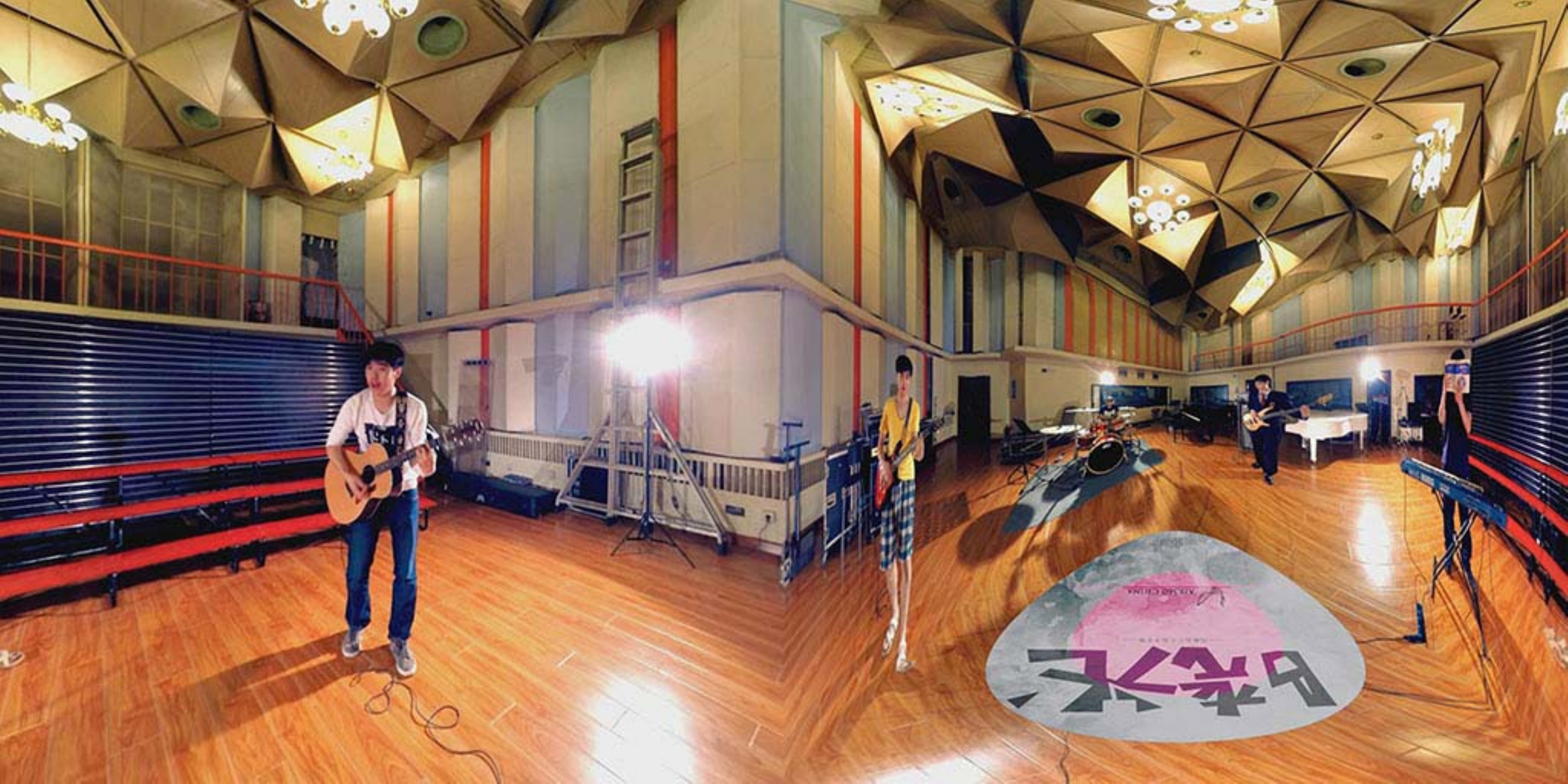}
}
\hspace{-1em}
}
\end{center}
\vspace{-1em}
\caption{Examples of different map projection types for one omnidirectional video frame.}
\label{fig:proj}
\end{figure*}
\section{Dataset Establishment}\label{sec:dataset}
In this section, we discuss how to establish the VQA-ODV dataset, from the aspects of omnidirectional sequences, subjective data collection and data formats. Our VQA-ODV dataset is available online at \url{https://github.com/Archer-Tatsu/VQA-ODV}.
\subsection{Omnidirectional sequences}
\textbf{Reference sequences}. Our dataset has in total 600 omnidirectional video sequences, of which 60 are reference sequences in diverse content. Among these 60 sequences, 12 sequences are in raw format \cite{vrsequences}, and others are downloaded from YouTube Virtual Reality Channel, the bitrates of which are more than 15 Mbps. As such, high quality can be ensured for the reference sequences. The reference sequences contain a wide range of content categories, such as scenery, computer graphics (CG), show and sports. The resolution of all reference sequences covers from 4K ($3840\times1920$ pixels) to 8K ($7680\times3840$ pixels). Additionally, the reference sequences are all under equirectangular projection (ERP). Then, the original video sequences are cut to make the duration of the reference sequences vary from 10 to 23 seconds under frame rate between 24-30 frames per second (fps).
We equally divide all reference sequences into 10 groups for facilitating subjective data collection.
To guarantee the diversity of sequences in each group, both the resolution and content categories of the reference sequences within a group are various. Figure \ref{fig:screen} shows all the reference sequences in our VQA-ODV dataset.

\textbf{Impaired sequences}. In our VQA-ODV dataset, we mainly take two kinds of impairment into account: compression and map projection. The former is ubiquitous to all types of encoded video, while the latter is a unique characteristic of omnidirectional video. In total, 3 compression levels and 3 kinds of projection are considered, and thus 9 different impaired sequences correspond to each reference sequence.
Specifically, we use H.265 \cite{Li2017Optimal, 8336889}, a state-of-the-art video compression standard, to compress the reference sequences. For each sequence, three test points are obtained with quantization parameter (QP) $=$ 27, 37 and 42 \cite{Zhou2017Complexity}. Accordingly, the bitrates are at high, medium and low levels \cite{Tao2015Improving}. Consequently, the subjective quality of the encoded sequences spans a large range.

We consider 3 projection types: ERP, reshaped cubemap projection (RCMP) and truncated square pyramid projection (TSP) \cite{auwera2016ahg8}. Examples of these projection types are shown in Figure \ref{fig:proj}. ERP is commonly used because of its simplicity. Additionally, the picture in an ERP frame is continuous everywhere. However, ERP has several drawbacks in the aspects of encoding efficiency, geometry distortion, etc. RCMP \cite{ng2005data} projects the spherical picture of an omnidirectional frame onto the six faces of its concentric cube (in a $3\times2$ face configuration). RCMP has no geometry distortion within the faces \cite{kuzyakov2016next}, but it incurs discontinuity across faces. In TSP, the front face takes up half area of the picture without any geometry distortion. However, other faces of TSP take up the remaining half picture with significant geometry distortion. Thus, the front face in TSP is much more important than other faces. In addition, there exist sharp edges in pictures under both RCMP and TSP projections \cite{sreedhar2016viewport}.

Each reference sequence is first converted to other map projections with Samsung 360tools\footnote{ \url{https://github.com/Samsung/360tools}}, and then compressed at different QPs using libx265 embedded in FFmpeg\footnote{ \url{http://www.ffmpeg.org/}}. Finally, 540 impaired omnidirectional video sequences are obtained at different bitrate levels and projections.
\subsection{Subjective data collection}
\begin{figure*}[!tb]
\begin{center}
\includegraphics[width=1\textwidth]{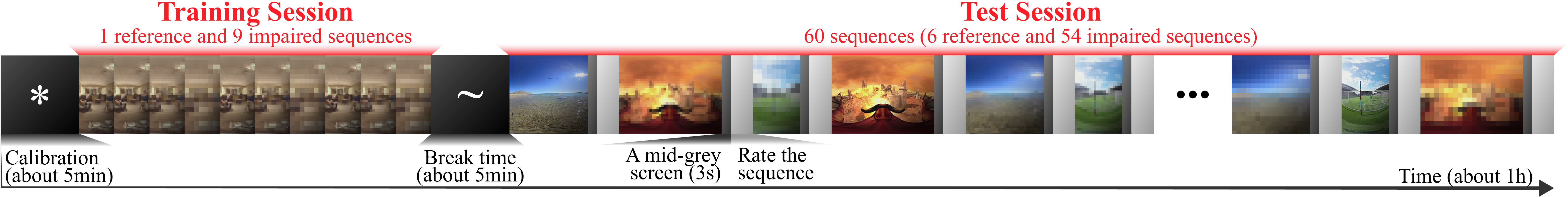}
\end{center}
\caption{Procedure of the experiment to rate the subjective quality scores of omnidirectional sequences. }
\label{fig:proc}
\end{figure*}
\begin{figure}[!tb]
\begin{center}
\includegraphics[width=1\linewidth]{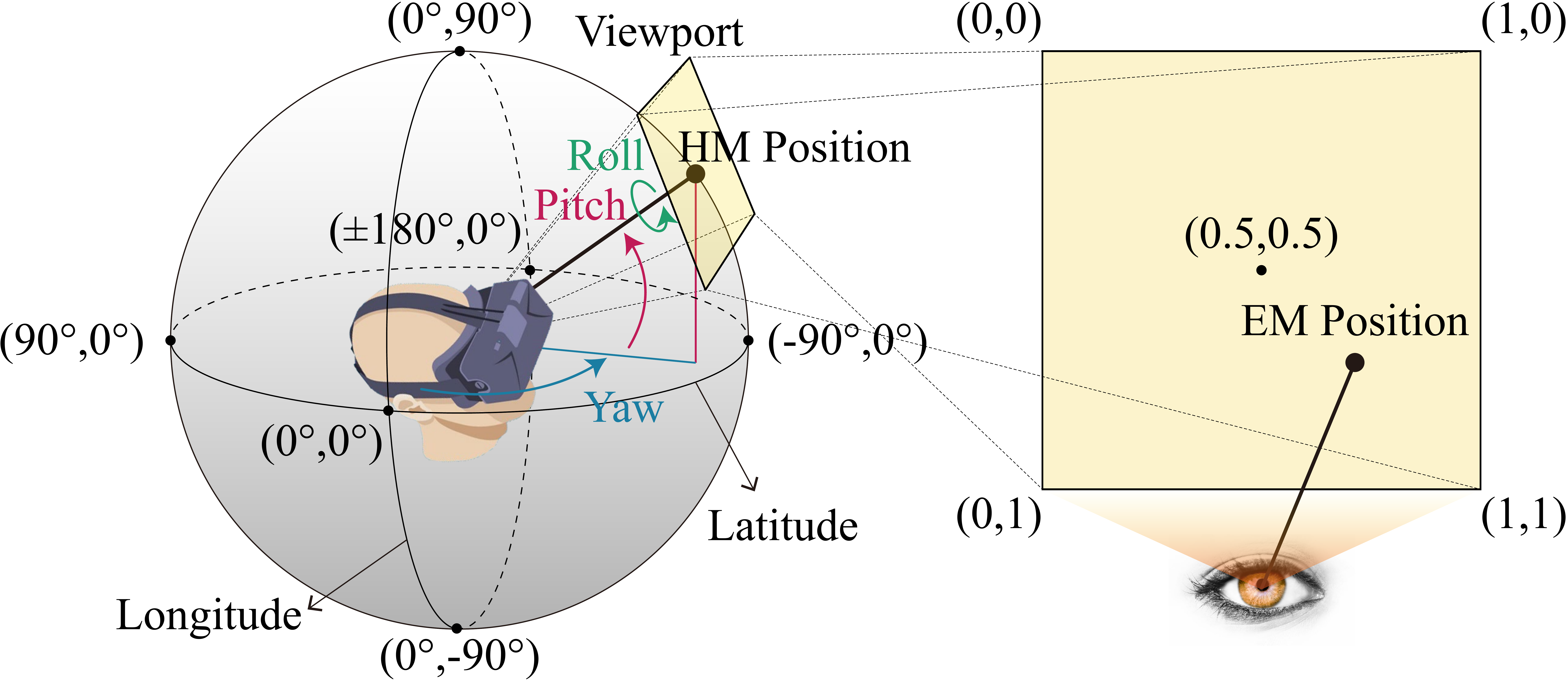}
\end{center}
\caption{An illustration of HM and EM. Left: HM in the sphere. The latitude and longitude of the HM position, i.e., the center of the viewport, are only dependent on the angle of pitch and yaw, respectively. The angle of roll only decides the posture of the viewport around its center. Right: the EM position in the viewport.}
\vspace{-0.5em}
\label{fig:coord}
\end{figure}
\textbf{Hardware and software}. In our experiment, HTC Vive is used as the HMD, connected to a high-performance computer. Additionally, an eye-tracking module, aGlass DKI\footnote{aGlass is able to provide low latency ($\ge\!120$Hz), high precision ($\le\!0.5^{\circ}$) and full viewport ($\ge\!110^{\circ}$), for eye tracking. See \url{http://www.aglass.com/?lang=en} for more details about this device. }, is embedded in HTC Vive to capture the eye-tracking data of subjects. We develop a graphic user interface (GUI) to control the experiment procedure, and we also develop a program to capture and save HM and EM data at a specific frequency. The frequency is set to be 2 times of the frame rate of the sequences, meaning that there are two sample points at one frame. A software, Virtual Desktop, is used in our experiment, not only as an omnidirectional video player but also to display our GUI in the HMD.

\textbf{Experiment procedure}. We follow the general settings and procedure proposed in \cite{xu2017subjective} to conduct the experiment. The procedure is illustrated in Figure \ref{fig:proc}. Generally speaking, the experiment procedure is composed of two sessions: the training and test sessions. Before the training session, the HMD and eye-tracking module are calibrated for each subject.
In the training session, the subject is told about the goal of our experiment. Then, the subject needs to watch the training sequences, in order to be familiar with omnidirectional video and its quality. Afterwards, there is a short break for further communication before entering the test session.
In the test session, we use a single stimulus paradigm so that the sequences are displayed in a random order with no direct comparison. After each sequence is displayed, there follows a 3-second mid-grey screen.

\textbf{Data collection}. There are two kinds of data to be collected in the experiment: (1) The raw subjective quality scores of the sequences; (2) The HM and EM data of subjects. In the test session, after watching each sequence, subjects are required to rate its quality with our GUI, as shown in Figure \ref{fig:proc}. A continuous quality scale is adopted with a range of 0 to 100, in which a large score means high quality. During the test session, the HM and EM data capture program starts simultaneously whenever a sequence starts being played. As a result, the captured HM and EM data are aligned with the video frames. Finally, the HM and EM data can be collected in the test session.

\textbf{Subjects.}
The total number of subjects participating in our experiment is 221, consisting of 143 males and 78 females. The age of subjects ranges from 19 to 35. These subjects are all of normal or corrected-to-normal vision.
Since the video sequences are divided into 10 groups, the subjects are also divided into 10 groups, such that each subject only watches one group of omnidirectional sequences for avoiding eye fatigue. After the experiment, the subject rejection \cite{series2012methodology} is applied according to the rating score from subjects. It is guaranteed that after the rejection, there are still no less than 20 valid subjects in each group.

\subsection{Subjective data formats}\label{sec:format}
\textbf{Subjective score data}. Both the mean opinion score (MOS) and DMOS are provided in our dataset. The MOS value of sequence $j$ can be calculated by
\begin{equation}
\mathrm{MOS}_j = \frac{1}{I_j}\sum_{i=1}^{I_j}S_{ij}\mbox{,}
\end{equation}
where $S_{ij}$ is the raw score that subject $i$ assigns to sequence $j$; $I_j$ is the number of valid subjects viewing sequence $j$. Refer to \cite{seshadrinathan2010study} for the calculation of DMOS, which qualifies the subjective quality degradation of impaired video. Note that the MOS scores of reference sequences have actual values, while their DMOS values are always 0.

\textbf{Content and formats}. The HM and EM data of a subject at one omnidirectional sequence are represented in a vector:
\\
\texttt{
[Timestamp HM\_pitch HM\_yaw HM\_roll EM\_x EM\_y EM\_flag].
}

\vspace{1em}
The content of the above vector is discussed in the following:
\begin{itemize}
  \item Timestamp. The interval time between two adjacent sample points is recorded and represented in milliseconds.
  \item HM data. There are 3 elements representing the 3 Euler angles of HM, which are the angles of pitch, yaw and roll, respectively. As shown in Figure \ref{fig:coord}, the position of the viewport only depends on the angles of pitch and yaw. This means that the second and third elements of the vector are equivalent to the latitude and longitude in Figure \ref{fig:coord}.
  \item EM data. There are 2 elements (\texttt{x} and \texttt{y}) representing the horizontal and vertical positions of EM within the viewport, as shown in Figure \ref{fig:coord}. The values of these elements are normalized, both falling in the range of $\left[0,1\right]$.
  \item Validity flag of EM data. The last element of the vector reflects whether the EM data of the corresponding sample point are valid (=1 for validness and =0 for invalidness). It is because the EM data cannot always be captured, e.g., eye blink may lead to invalid EM data. Note that the HM data are all valid due to the operation principle of the HMD.
\end{itemize}
\begin{table}[!tb]
  \centering
  \caption{Mean values and standard deviations of DMOS under different projections and bitrate levels.}
    \begin{tabular}{cccc}
    \toprule
    Projections & High  & Medium & Low \\
    \midrule
    ERP   & 39.05 $\pm$ \textbf{3.60} & 50.40 $\pm$ 8.27 & 62.13 $\pm$ \textbf{8.64} \\
    RCMP  & 39.04 $\pm$ 3.69 & 51.53 $\pm$ 8.43 & 64.50 $\pm$ 9.09 \\
    TSP   & \textbf{38.29} $\pm$ 4.01 & \textbf{47.23} $\pm$ \textbf{7.01} & \textbf{57.82} $\pm$ 8.95 \\
    \bottomrule
    \end{tabular}%
  \label{tab:msDMOS}%
\end{table}%
\begin{figure}[!tb]
\begin{center}
\resizebox{1\linewidth}{!}{
\subfigure[Best DMOS scores]{
  \label{fig:bwbar:best}
  \includegraphics[width=0.5\linewidth]{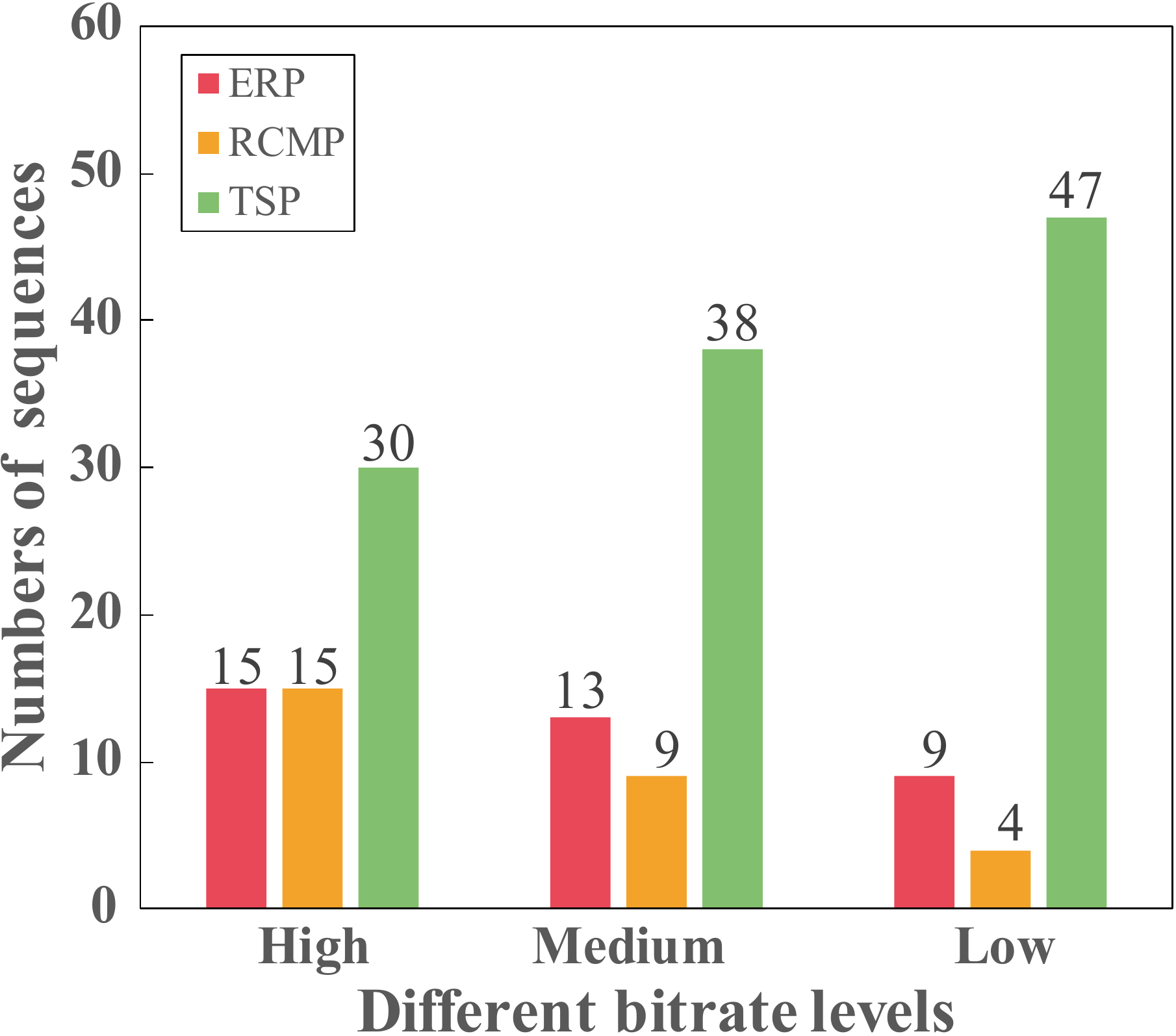}
}
\subfigure[Worst DMOS scores]{
  \label{fig:bwbar:worst}
  \includegraphics[width=0.5\linewidth]{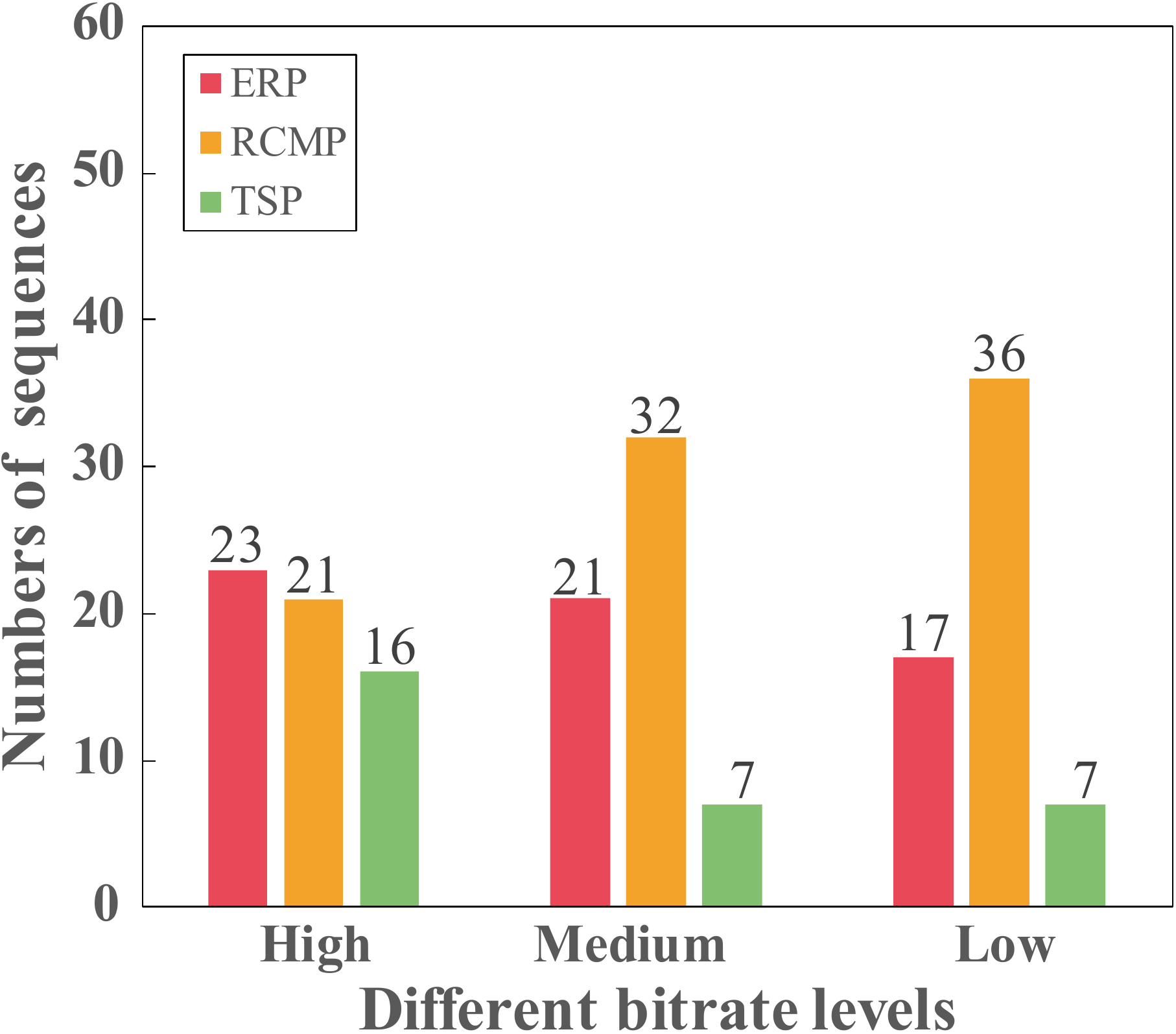}
}
}
\caption{Numbers of sequences achieving the best and worst DMOS scores under different projections, for each bitrate level. }
\label{fig:bwbar}
\end{center}
\end{figure}

\section{Data Analysis}
In this section, we analyze our VQA-ODV dataset in the following three aspects.
\begin{table*}[!tb]
  \centering
  \caption{Performance of the SSIM metric and the PSNR related metrics. Data in the ``Mean" column are the results averaged over all 10 groups of VQA-ODV. ``All Data" column means the correlation results over all 540 impaired sequences of the VQA-ODV dataset, reflecting the performance in the whole dataset.}
  \begin{tabular}{|c|cc|cc|cc|cc|cc|cc|}
    \hline
    \multirow{3}{*}{Metrics} & \multicolumn{6}{c|}{Without HM and EM} & \multicolumn{6}{c|}{With HM and EM} \\
    \cline{2-13}
     & \multicolumn{2}{c|}{SSIM} & \multicolumn{2}{c|}{PSNR} & \multicolumn{2}{c|}{S-PSNR} & \multicolumn{2}{c|}{$\text{PSNR}_{\text{O-HM}}$} & \multicolumn{2}{c|}{$\text{PSNR}_{\text{I-HM}}$} & \multicolumn{2}{c|}{$\text{PSNR}_{\text{I-EM}}$} \\
    \cline{2-13}
      & Mean  & All Data & Mean  & All Data & Mean  & All Data & Mean  & All Data & Mean  & All Data & Mean  & All Data \\
    \hline
    PCC & 0.60 & 0.44 & 0.56 & 0.49 & 0.61 & 0.55 & 0.66 & 0.60 & 0.67 & 0.61 & \textbf{0.70} & \textbf{0.67} \\
    SRCC & 0.62 & 0.49 & 0.55 & 0.51 & 0.62 & 0.58 & 0.67 & 0.62 & 0.67 & 0.63 & \textbf{0.70} & \textbf{0.69} \\
    RMSE & 9.25 & 10.63 & 9.70 & 10.36 & 9.29 & 9.95 & 8.65 & 9.51 & 8.48 & 9.43 & \textbf{8.09} & \textbf{8.84} \\
    MAE & 7.27 & 8.31 & 7.77 & 8.11 & 7.36 & 7.75 & 6.87 & 7.46 & 6.73 & 7.38 & \textbf{6.30} & \textbf{6.90} \\
    \hline
    \end{tabular}%
  \label{tab:corr}%
\end{table*}%
\subsection{Analysis on VQA Results}\label{sec:vqa}
First, we focus on the subjective VQA results at different impairment types. Table \ref{tab:msDMOS} shows the mean values and standard deviations of DMOS under different projections and bitrate levels, which are obtained over all 10 groups of our VQA-ODV dataset.
It is obvious that the sequences with higher bitrates are of better subjective quality, even at different projections.
Thus, compared to projection, bitrate has much more impact on subjective quality of impaired omnidirectional video.
Figure \ref{fig:bwbar} further plots the numbers of sequences that have the best and worst DMOS scores under different projections, for each bitrate level.
As can be seen in Table \ref{tab:msDMOS} and Figure \ref{fig:bwbar}, at the same bitrate, the TSP projection can yield slightly better subjective quality than other two projections.
We can further see that the impact of projection is insignificant at high bitrates.

Next, we measure the objective VQA results of all impaired sequences, in terms of PSNR, S-PSNR \cite{yu2015framework} and structural similarity (SSIM) index \cite{wang2004image}. We perform a non-linear regression on these objective VQA scores using a logistic function following \cite{seshadrinathan2010study}.
Then, we evaluate the performance of these objective video quality metrics by calculating the correlation between the fitted objective scores and the corresponding subjective quality of DMOS. Here, the correlation is evaluated by Spearman's rank correlation coefficient (SRCC), Pearson correlation coefficient (PCC), root-mean-square error (RMSE) and mean absolute error (MAE).
Note that the large values of SRCC and PCC, or small values of RMSE and MAE, mean high correlation between objective and subjective VQA.
Table \ref{tab:corr} reports the correlation results on each group and the entire dataset.
We can see from Table \ref{tab:corr} that S-PSNR, which is developed for assessing omnidirectional video, performs better than the traditional PSNR and SSIM metrics.

Table \ref{tab:tradition} summarizes the correlation results of the traditional PSNR and SSIM metrics with subjective quality over 2D video, which were reported in the existing literature.
We can further see from Tables \ref{tab:corr} and \ref{tab:tradition} that the traditional VQA metrics are not effective in assessing the quality of omnidirectional video, especially compared to that of 2D video.
It is intuitive that human behavior plays an important role in determining the visual quality of omnidirectional video, since human is the ultimate end of omnidirectional video. In the following, we thoroughly analyze the human viewing behavior on omnidirectional video.
\begin{figure}[!tb]
\begin{center}
\resizebox{1\linewidth}{!}{
\subfigure[PCC for HM weight maps]{
  \label{fig:human:cchm}
  \includegraphics[width=0.5\linewidth]{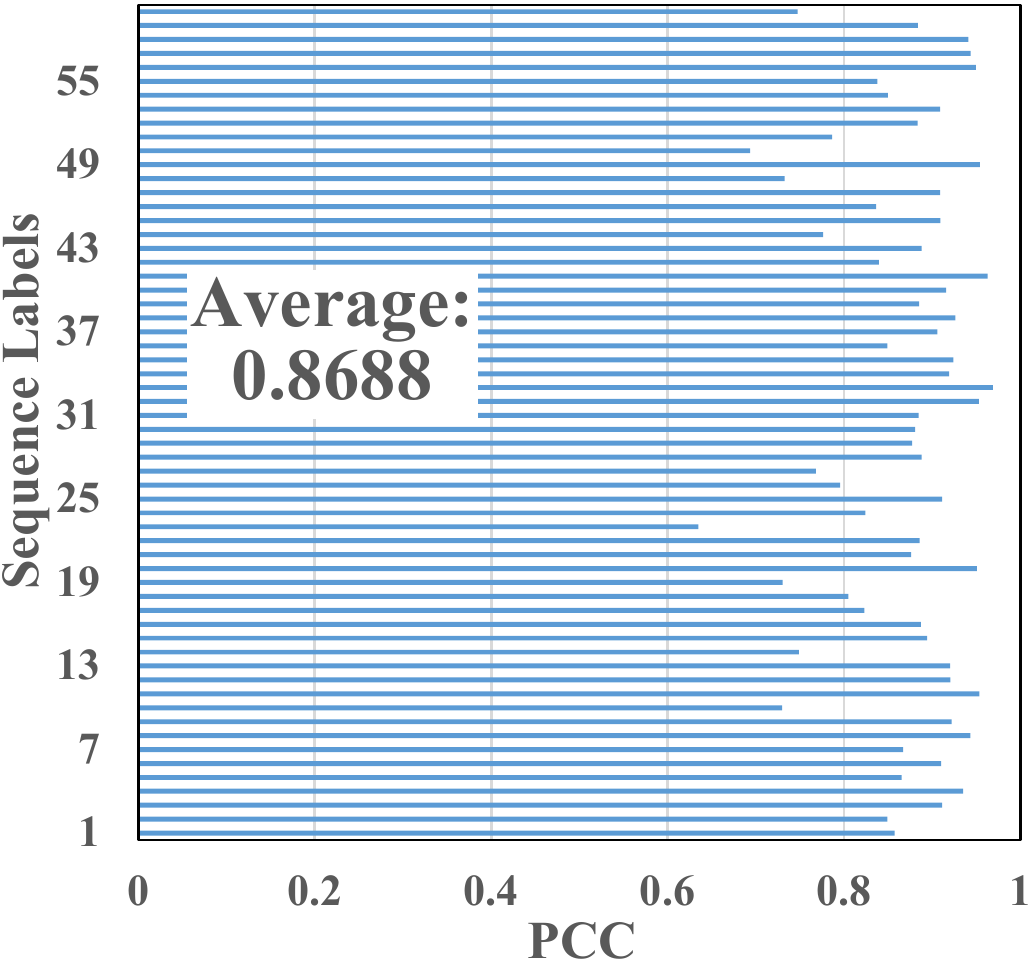}
}

\subfigure[PCC for EM weight maps]{
  \label{fig:human:ccem}
  \includegraphics[width=0.5\linewidth]{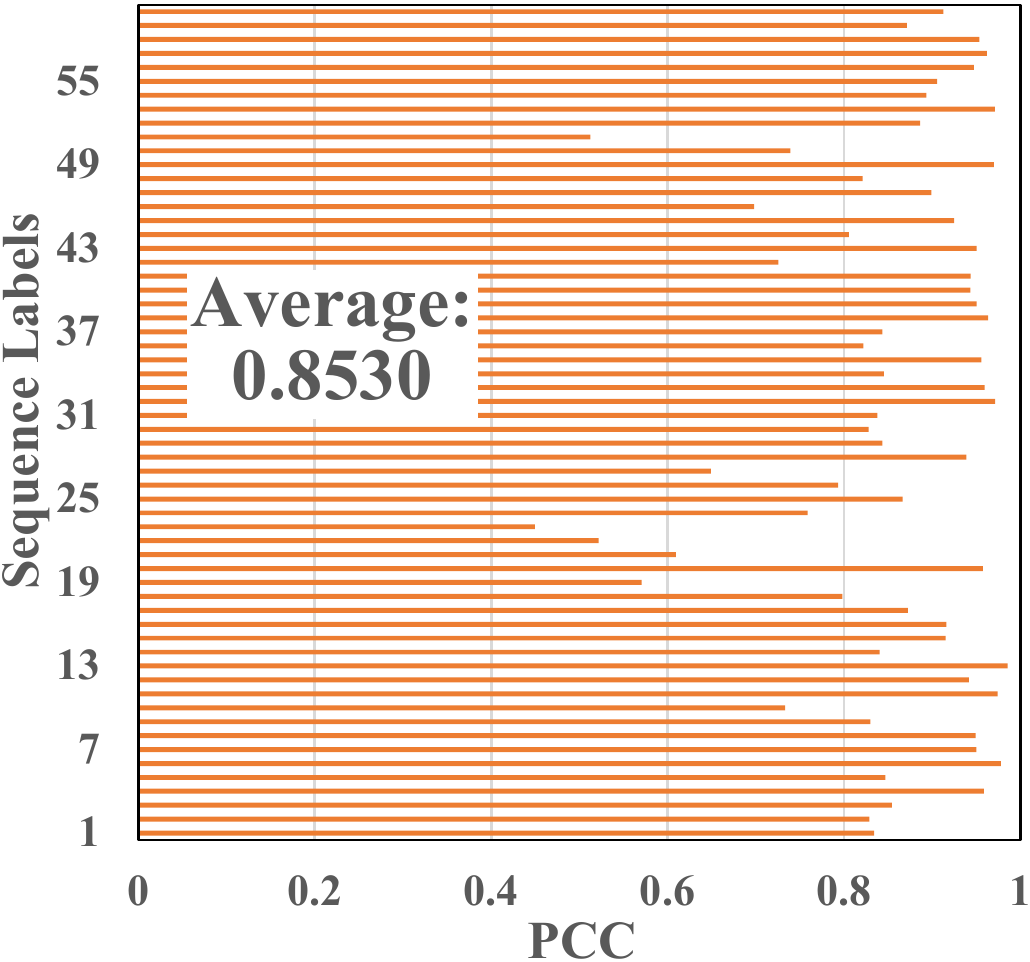}
}
}
\end{center}
\vspace{-0.5em}
\caption{PCC results of HM and EM weight maps between two sub-groups for each of the 60 reference sequences in our VQA-ODV dataset.}
\label{fig:human}
\vspace{-1em}
\end{figure}
\begin{figure}[!tb]
\begin{center}
  \includegraphics[width=1\linewidth]{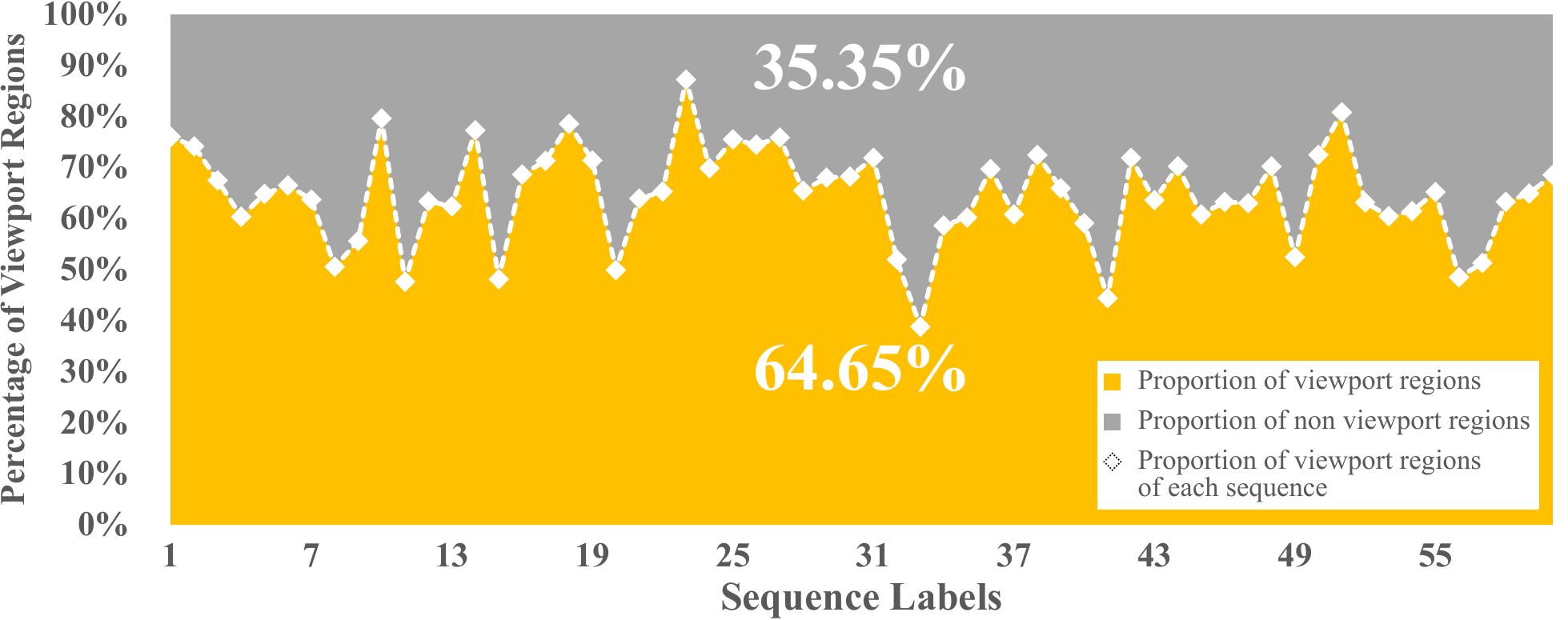}
\end{center}
\caption{Percentage of viewport regions, for each of 60 reference sequences in our VQA-ODV dataset. }
\vspace{-1.5em}
\label{fig:area}
\end{figure}
\subsection{Analysis on Human Behavior}
\begin{table}[!tb]
  \centering
  \caption{Performance of PSNR and SSIM on 2D video.}
    \resizebox{1\linewidth}{!}{
    \begin{tabular}{cccccccc}
    \toprule
    & Metrics & Our & \tabincell{c}{St\"{u}tz\\et al. \cite{stutz2010subjective}} & \tabincell{c}{Liotta\\et al. \cite{liotta2013instantaneous}} & \tabincell{c}{Rahman\\et al. \cite{rahman2017reduced}} & \tabincell{c}{Dostal\\et al. \cite{dostal2012hlfsim}} & \tabincell{c}{Park\\et al. \cite{park2012new}} \\
    \midrule
    \multirow{3}[2]{*}{PSNR} & PCC & 0.49 & 0.91 & 0.93 & 0.90 & 0.80 & - \\
      & SRCC & 0.51 & 0.91 & - & 0.87 & 0.82 & - \\
      & RMSE & 10.36 & 0.56 & 0.33 & - & - & - \\
    \midrule
    \multirow{3}[2]{*}{SSIM} & PCC & 0.44 & 0.84 & 0.97 & 0.82 & 0.72 & 0.93 \\
      & SRCC & 0.49 & 0.86 & - & 0.76 & 0.85 & 0.94 \\
      & RMSE & 10.63 & 0.73 & 0.17 & - & - & - \\
    \bottomrule
    \end{tabular}
    }%
  \label{tab:tradition}%
  \vspace{-1em}
\end{table}%
\begin{figure*}[!tb]
\begin{center}
\resizebox{\textwidth}{!}{
\hspace{-1em}
\subfigure[I-HM]{
  \label{fig:weight:hmone}
  \includegraphics[width=0.25\textwidth]{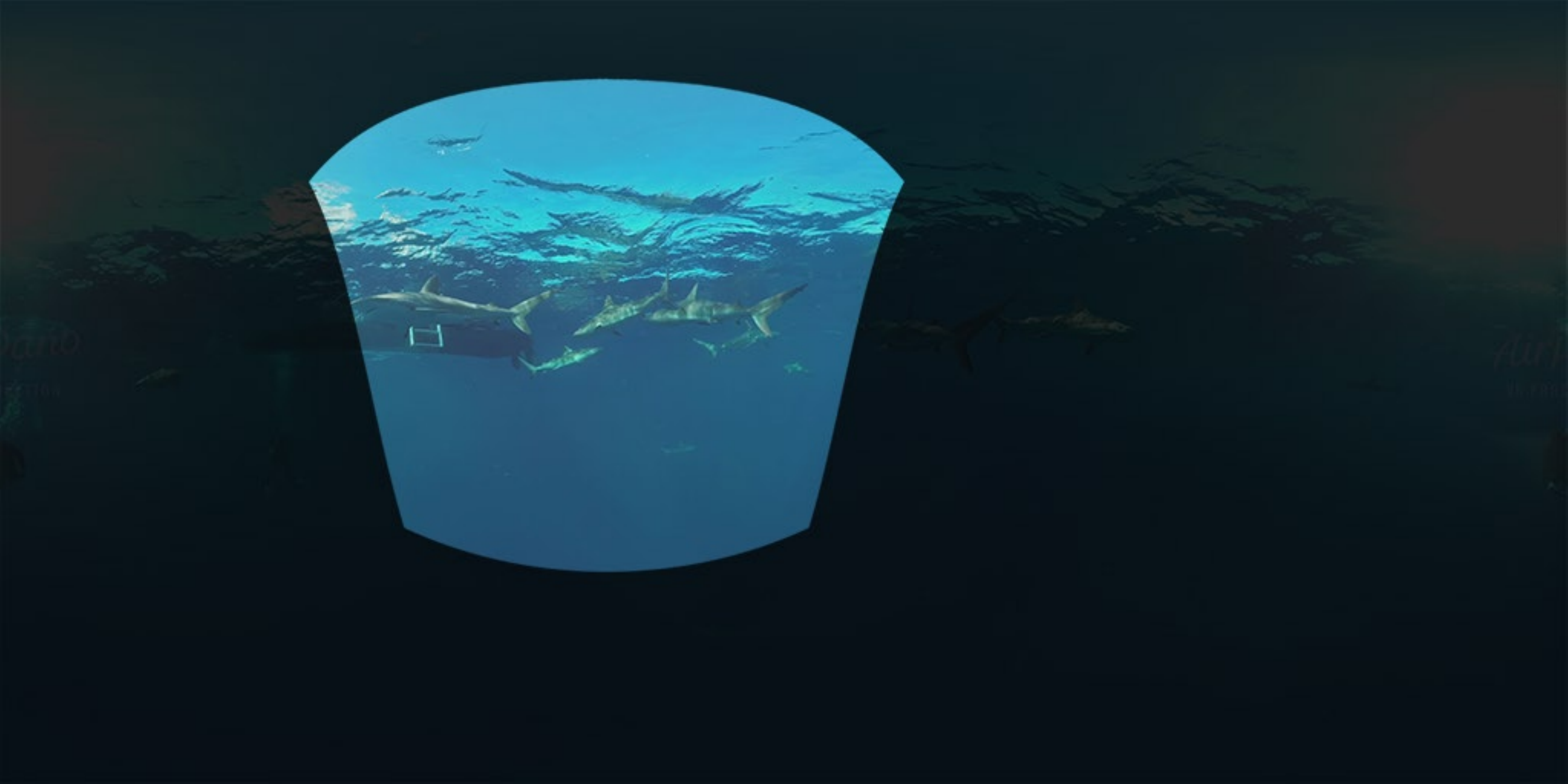}
}
\subfigure[O-HM]{
  \label{fig:weight:hmall}
  \includegraphics[width=0.25\textwidth]{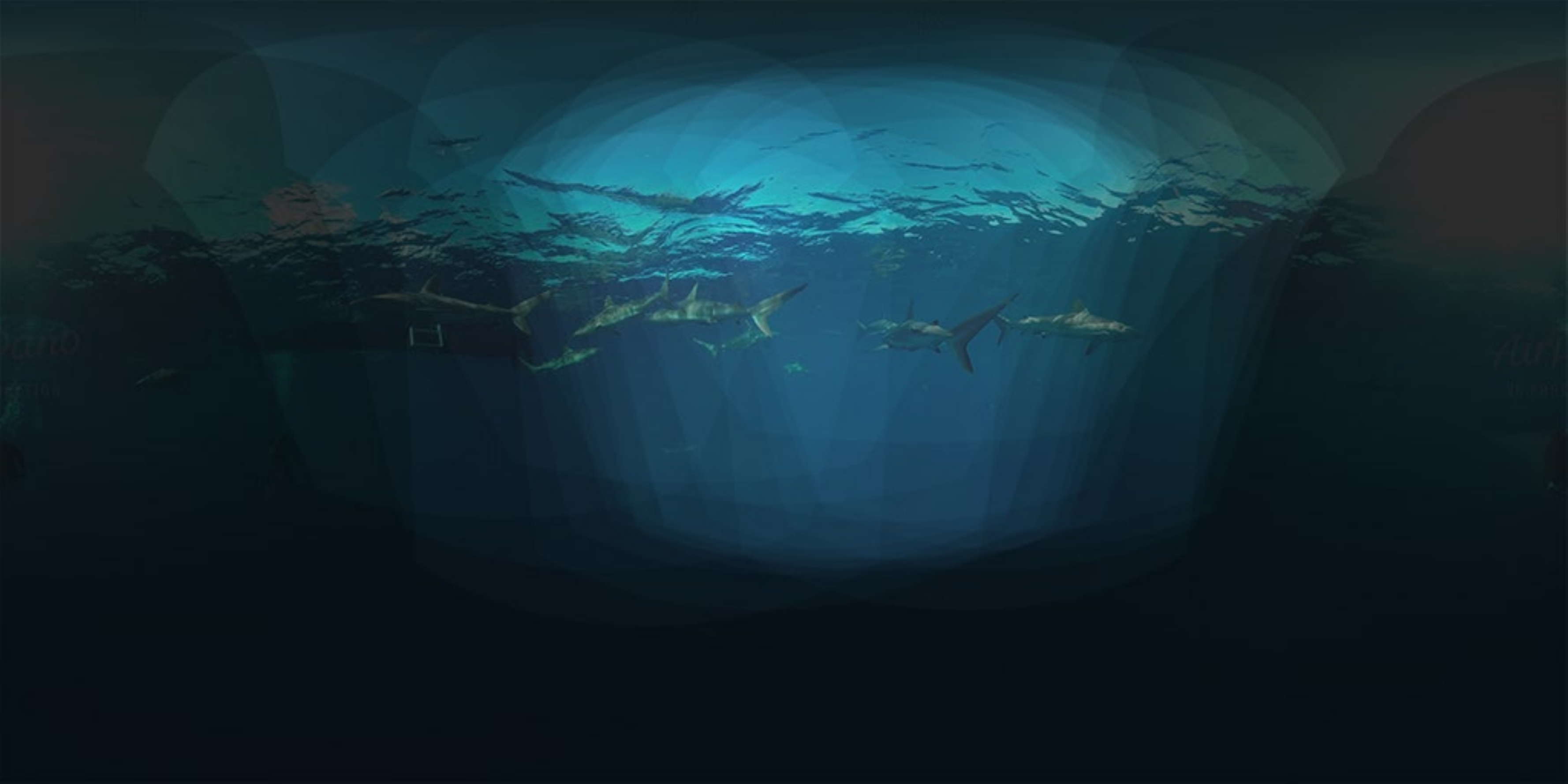}
}
\subfigure[I-EM]{
  \label{fig:weight:emmap}
  \includegraphics[width=0.25\textwidth]{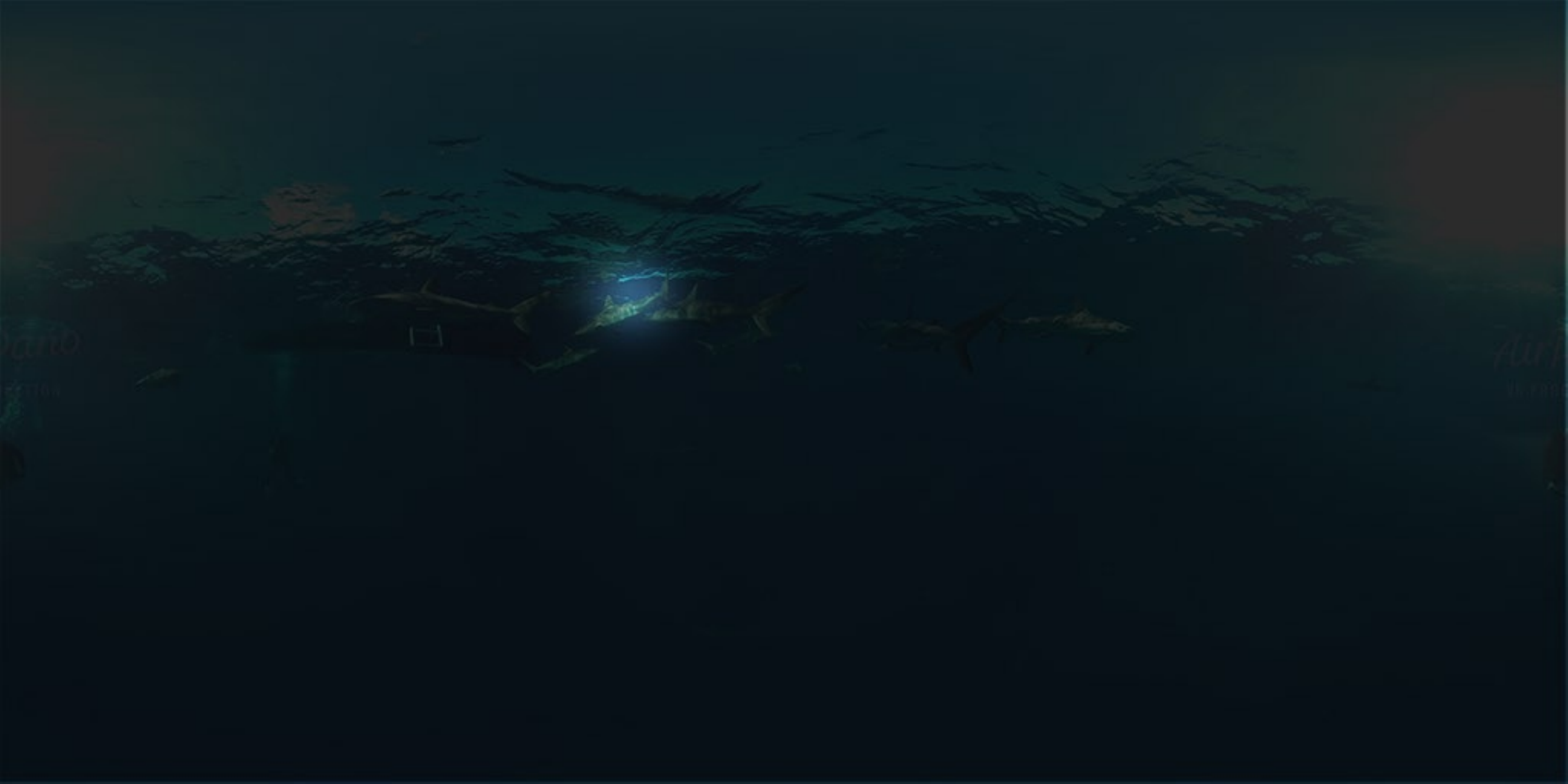}
}
\hspace{-1em}
}
\end{center}
\vspace{-1em}
\caption{Examples of one omnidirectional frame masked by the corresponding weight maps generated from HM and EM data. }
\label{fig:weight}
\end{figure*}
Here, we investigate the human behavior on viewing omnidirectional video via analyzing the HM and EM data over our VQA-ODV dataset. Specifically, we measure the consistency of HM and EM data across different subjects. To this end, we randomly and equally divide all subjects of the VQA-ODV dataset into two non-overlapping sub-groups. Then, the HM and EM weight maps of these two sub-groups are generated for all frames of the reference sequences. Note that the generation of HM and EM weight maps is based on the method of Section \ref{sec:corr}. Figure \ref{fig:human:cchm} shows the PCC results of HM weight maps between two sub-groups of subjects for all 60 sequences. We can see from this figure that there exists high HM consistency across different subjects, as the average PCC value \footnote{We find that the PCC value of randomly generated HM maps (in uniform distribution) is $4\times10^{-4}$, as the random baseline.} is 0.8688. Similar consistency exists for the EM weight maps of different subjects, which can be found from Figure \ref{fig:human:ccem}.

Furthermore, we find that each omnidirectional frame has some perceptual redundancy, as the viewport regions of all subjects cannot cover the whole $360 \times 180^\circ$ omnidirectional range. Figure \ref{fig:area} plots the percentage of all subjects' viewport regions to whole $360 \times 180^\circ$ omnidirectional region, which is averaged over all frames for each reference sequence in our VQA-ODV dataset. As shown in this figure, less than 65\% region of omnidirectional frames is viewed by all subjects. However, such a partial region decides the quality perceived by human. Thus, it is reasonable to take human behavior into consideration in VQA of omnidirectional video, which is to be discussed in the following.

\subsection{Impact of Human Behavior on VQA}\label{sec:corr}
To explore the impact of human behavior on VQA, we assign the perceptual weights in assessing PSNR of each impaired omnidirectional frame. The perceptual weights are based on the ground truth HM and EM data of human behavior. Here, we calculate 3 types of perceptual weights, called overall HM (O-HM), individual HM (I-HM) and individual EM (I-EM) weights.
Specifically, assume that the viewport of subject $i$ at one frame corresponds to a set of pixels denoted by $\mathbb{V}_i$.
Then, the I-HM weight map $w_i^{\mathrm{I-HM}}$ of subject $i$ can be obtained as follows,
\begin{equation}
w_i^{\text{I-HM}}(\mathbf{p})=\left\{
\begin{aligned}
1&, \mathbf{p}\in \mathbb{V}_i\\
0&, \mathbf{p}\in \text{others}
\end{aligned}\right.\text{,}
\end{equation}
where $\mathbf{p}$ is a pixel at the omnidirectional frame.
Figure \ref{fig:weight:hmone} shows an example of $w_i^{\text{I-HM}}$.
Subsequently, the O-HM weight map of all $I$ subjects for each omnidirectional frame can be calculated as
\begin{equation}
\label{eq:norm}
w^{\text{O-HM}}(\mathbf{p})=\frac{\sum_{i=1}^{I}  w_i^{\text{I-HM}}(\mathbf{p})}{\sum_{\mathbf{p}\in\mathbb{P}} \sum_{i=1}^{I} w_i^{\text{I-HM} }(\mathbf{p})}\text{.}
\end{equation}
In \eqref{eq:norm}, $\mathbb{P}$ denotes the set of all pixels at the omnidirectional frame.
Figure \ref{fig:weight:hmall} shows an example of the O-HM weight map.

Then, we turn to the calculation of the I-EM weight map. Given the EM position $\mathbf{e}_i$ of subject $i$, the I-EM weight map $w_i^{\mathrm{I-EM}}$ can be generated via Gaussian filtering in the viewport:
\begin{equation}
w_i^{\text{I-EM}}(\mathbf{p})=\left\{
\begin{aligned}
\exp\left(-\frac{\|\mathbf{e}_\mathbf{p}-\mathbf{e}_i\|_2^2}{2\sigma^2}\right)&, \mathbf{p}\in \mathbb{V}_i\\
0&, \mathbf{p}\in \text{others}
\end{aligned}\right.\text{,}
\end{equation}
where $\mathbf{e}_\mathbf{p}$ is the position of pixel $\mathbf{p}$ at the viewport, and $\sigma$ is the standard deviation of Gaussian distribution. Figure \ref{fig:weight:emmap} provides an example of the I-EM map.

Next, the error between the reference and impaired frames is weighted by $w_i^{\mathrm{I-HM}}$, $w^{\mathrm{O-HM}}$ and $w_i^{\mathrm{\text{I-EM}}}$ to obtain $\text{PSNR}_{\text{I-HM}}$, $\text{PSNR}_{\text{O-HM}}$ and $\text{PSNR}_{\text{I-EM}}$, respectively.
More specifically, $\text{PSNR}_{\text{I-HM}}$ and $\text{PSNR}_{\text{I-EM}}$ are calculated by averaging weighted PSNR over all subjects:
\begin{equation}
\text{PSNR}_{\text{I-EM}}=\frac{1}{I}\sum_{i=1}^{I}
10\log\frac{Y_{\mathrm{max}}^2 \cdot \sum_{\mathbf{p}\in\mathbb{P}}{w_i^{\text{I-EM}}(\mathbf{p})}}
{\sum_{\mathbf{p}\in\mathbb{P}}{\left(Y(\mathbf{p})\!-\!Y'(\mathbf{p})\right)^2\!\cdot\!w_i^{\text{I-EM}}(\mathbf{p})}}\text{,}
\end{equation}
\begin{equation}
\text{PSNR}_{\text{I-HM}}=\frac{1}{I}\sum_{s=1}^{I}
10\log\frac{Y_{\mathrm{max}}^2 \cdot \sum_{\mathbf{p}\in\mathbb{P}}{w_i^{\text{I-HM}}(\mathbf{p})}}
{\sum_{\mathbf{p}\in\mathbb{P}}{\left(Y(\mathbf{p})\!-\!Y'(\mathbf{p})\right)^2\!\cdot\!w_i^{\text{I-HM}}(\mathbf{p})}}\nonumber\text{,}
\end{equation}
where $Y(\mathbf{p})$ and $Y'(\mathbf{p})$ are intensities of pixel $\mathbf{p}$ in the reference and impaired omnidirectional frames, respectively. Additionally, $Y_{\mathrm{max}}$ is the maximum intensity value of the video sequences (=255 for 8-bit intensity).
Based the overall HM map, $\text{PSNR}_{\text{O-HM}}$ is obtained as follows,
\begin{equation}
\text{PSNR}_{\text{O-HM}}=10\log\frac{Y_{\mathrm{max}}^2}
{\sum_{\mathbf{p}\in\mathbb{P}}{\left(Y(\mathbf{p})\!-
\!Y'(\mathbf{p})\right)^2\!\cdot\!w^{\text{O-HM}}(\mathbf{p})}}.
\end{equation}

Finally, we measure the correlation between the above PSNR metrics (after non-linear regression) and DMOS values over our VQA-ODV dataset. The results are also reported in Table \ref{tab:corr}. As shown in this table, when incorporating the HM data of subjects, the performance of PSNR can be significantly improved. The EM maps are able to further improve the performance of PSNR over HM-weighted PSNR. This indicates that the HM and EM maps can be used to improve the effectiveness of objective VQA on omnidirectional video.
\section{Deep learning based VQA model}
\subsection{Method}
Recently, deep learning has witnessed a great success for objective VQA in 2D image and video \cite{kim2017deep,kim2017fully,vega2017deep,guan2017visual,hou2015blind}.
Unfortunately, to the best of our knowledge, there is no deep learning model for objective VQA on omnidirectional video.
Therefore, we propose a deep learning based VQA model for omnidirectional video, taking advantage of our large-scale dataset VQA-ODV.
More importantly, the weight maps of HM and EM are integrated into our deep learning model, which seamlessly bridges the gap between human behavior and VQA on omnidirectional video.
In the following, we present our deep learning model from the aspects of architecture and loss function.

\begin{figure}[!tb]
\begin{center}
\includegraphics[width=1\linewidth]{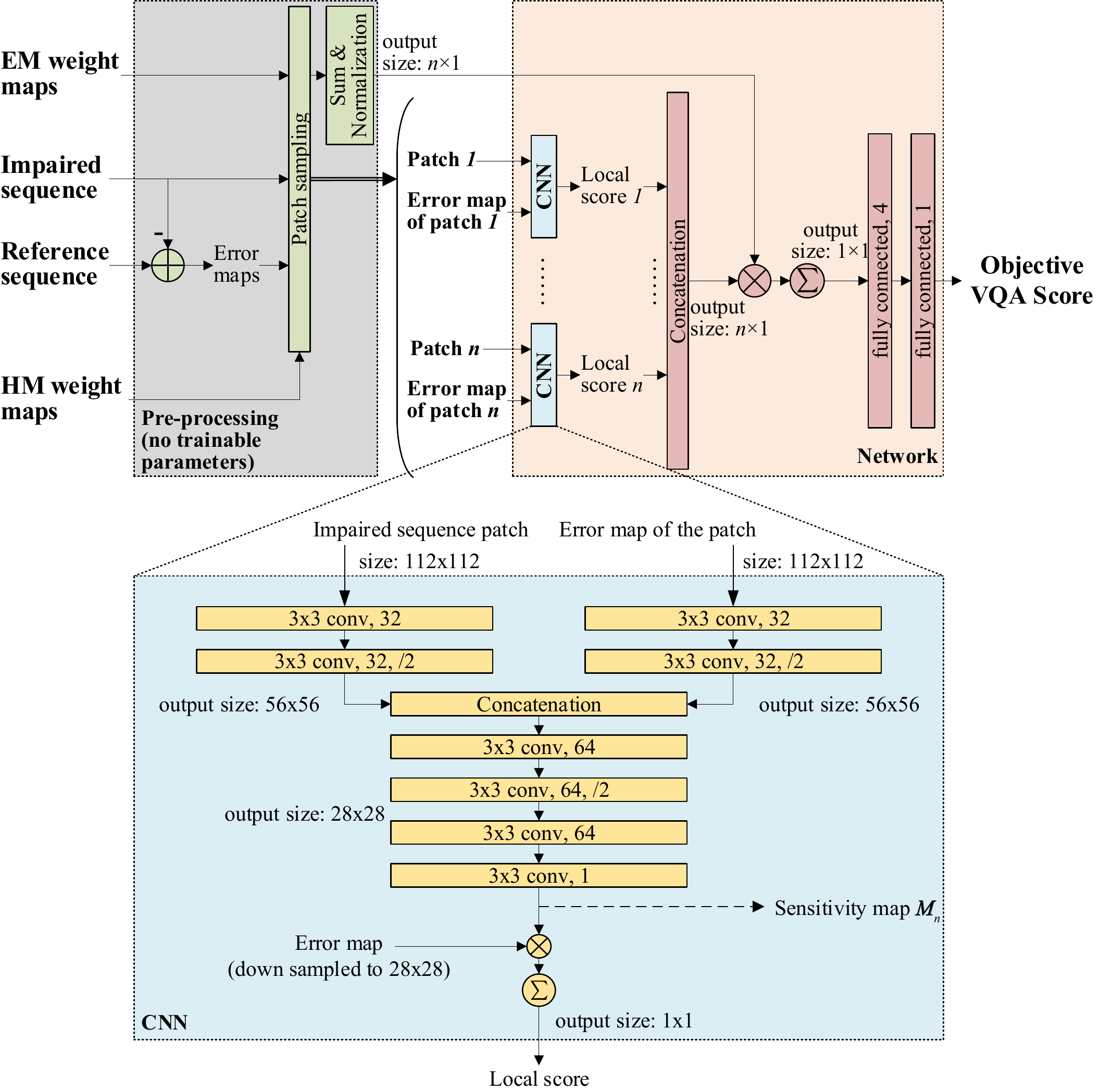}
\end{center}
\vspace{-1em}
\caption{The architecture of our VQA model.}
\vspace{-1em}
\label{fig:model}
\end{figure}

\textbf{Architecture.}  The architecture of our deep learning model is shown in Figure \ref{fig:model}. As shown in this figure, both impaired and reference sequences are input into our model, followed by a pre-processing step.
In the pre-processing step, the error maps between the impaired and reference sequences are generated. Additionally, given a patch and the HM or EM weight map of its corresponding frame, the values in the weight map that correspond to the pixels of the patch are summed to be one value, as the weight of the patch.
Then, $n$ impaired patches (size: 112 $\times$ 112) are sampled from each omnidirectional sequence with probabilities being the HM weights of the patches.
The error maps of $n$ impaired patches are also obtained. This way, our model only considers the content of human viewport in assessing visual quality of omnidirectional video.
Subsequently, each pair of the impaired patch and the corresponding error map is fed into a convolutional neural network (CNN) component, which is proposed in \cite{kim2017deep} for VQA of 2D images.
As a result, $n$ local VQA scores are obtained and then concatenated into a vector.
Corresponding to the VQA vector, an $n$-dimensional weight vector of the $n$ impaired patches is generated by dividing the EM weights of the sampled patches by their sum as normalization.
Subsequently, the $n$-dimensional VQA and weight vectors are with inner product, followed by two fully connected layers. As such, EM can be incorporated in our model.
Finally, the objective VQA score of the omnidirectional sequence is estimated.
\begin{figure*}[!tb]
\begin{center}
\hspace{-1em}
\subfigure[Our]{
  \label{fig:scatter:our}
  \includegraphics[width=0.19\textwidth]{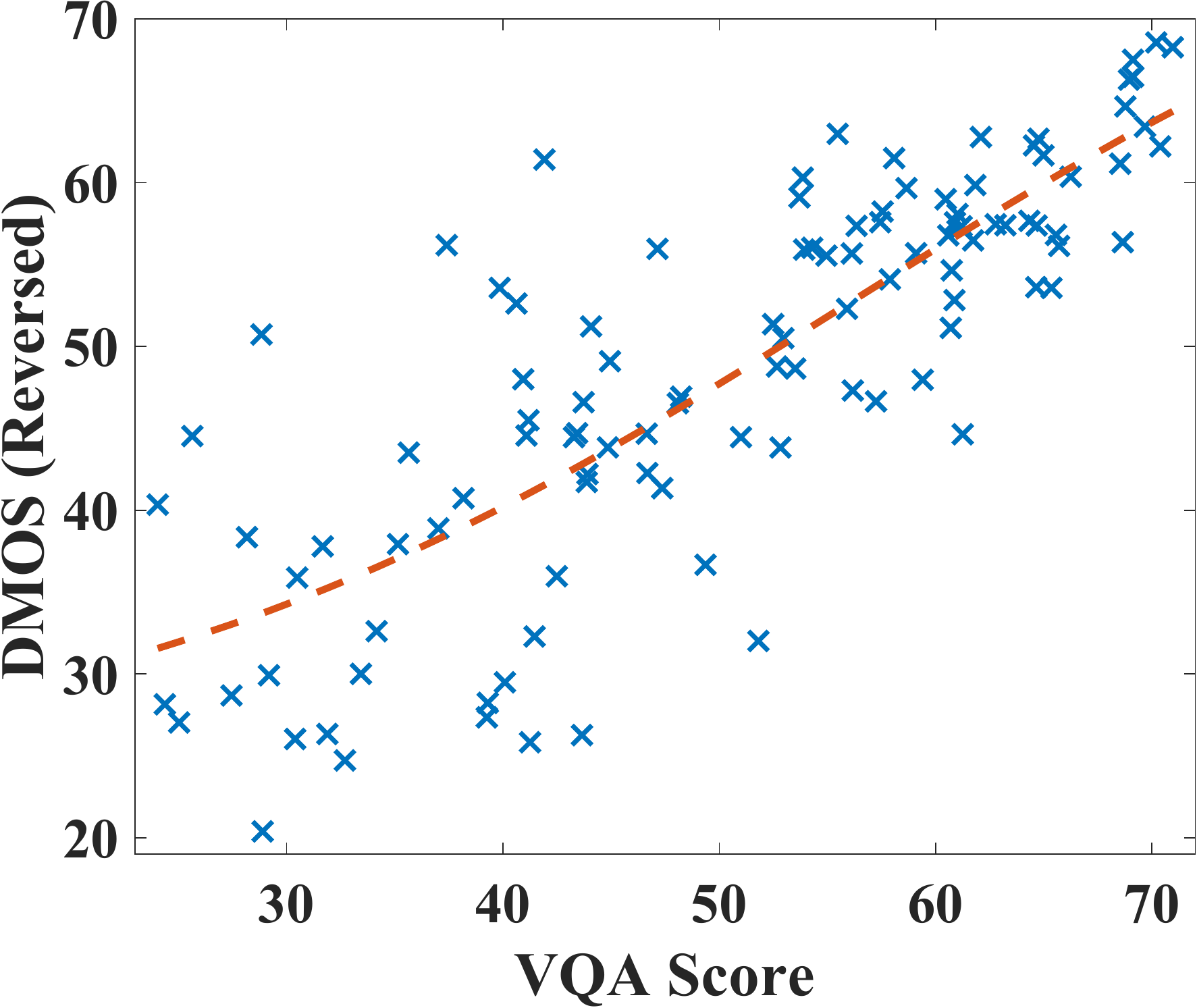}
}
\subfigure[DeepQA]{
  \label{fig:scatter:deepqa}
  \includegraphics[width=0.19\textwidth]{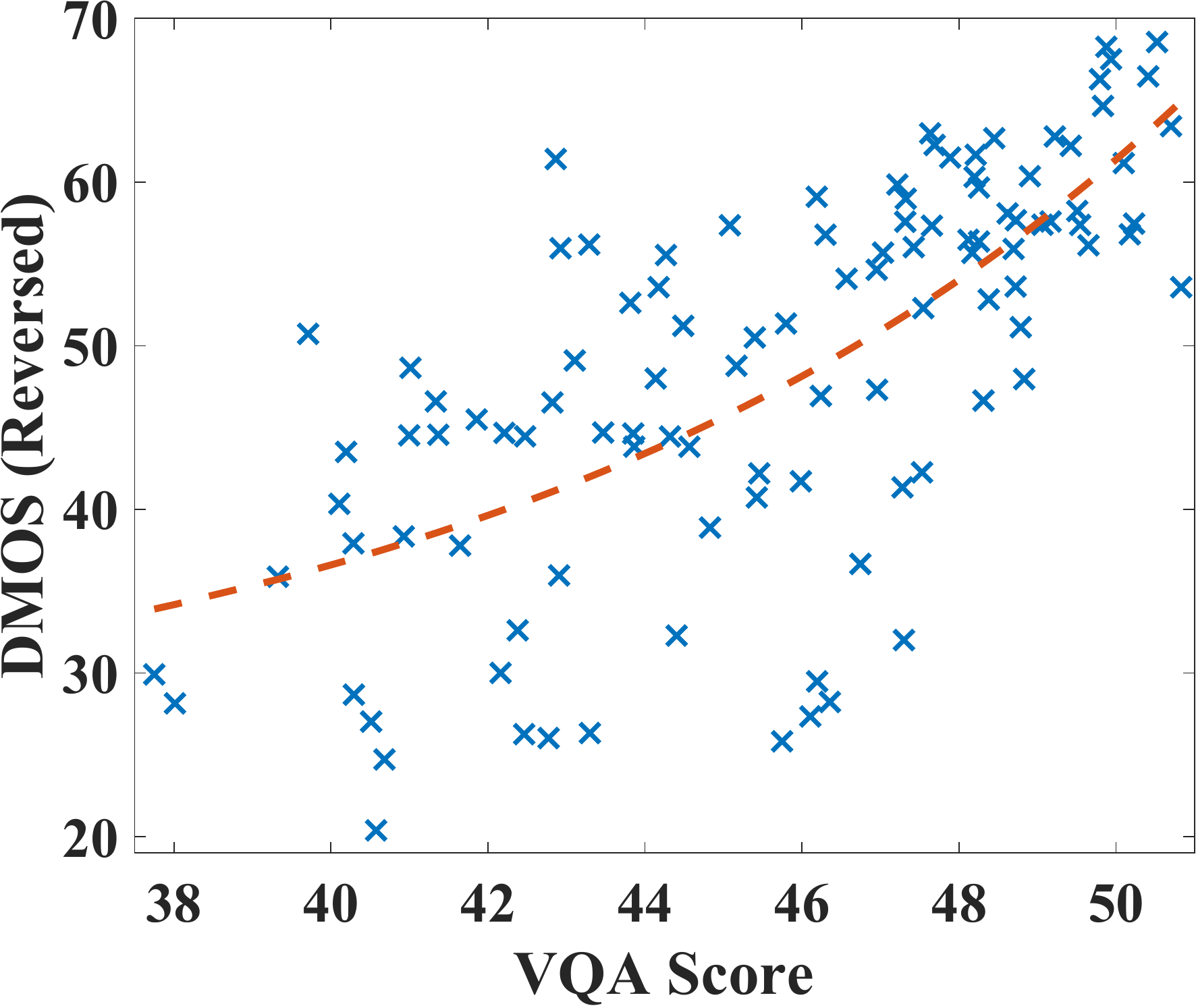}
}
\subfigure[S-PSNR]{
  \label{fig:scatter:spsnr}
  \includegraphics[width=0.19\textwidth]{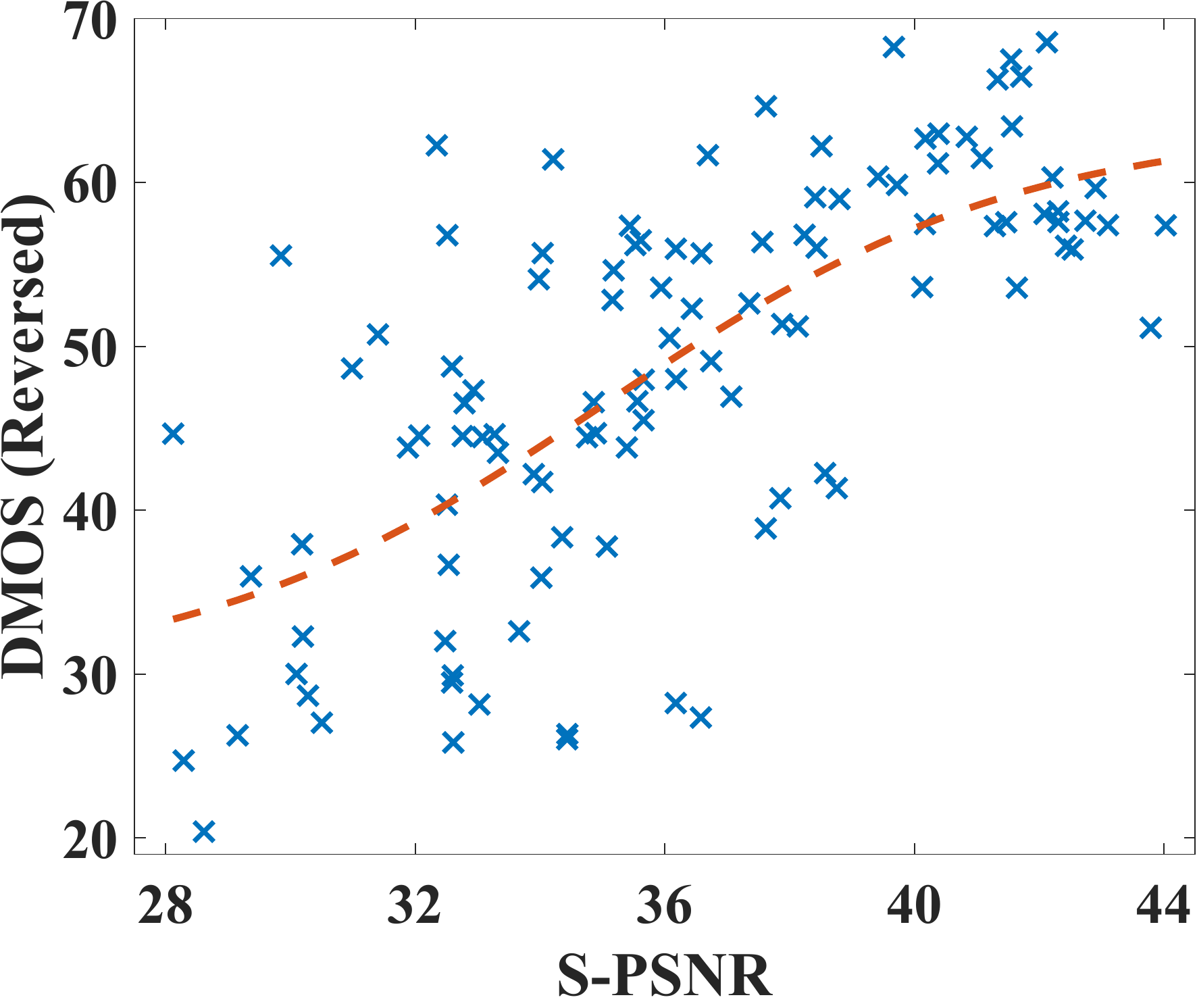}
}
\subfigure[CPP-PSNR]{
  \label{fig:scatter:cpppsnr}
  \includegraphics[width=0.19\textwidth]{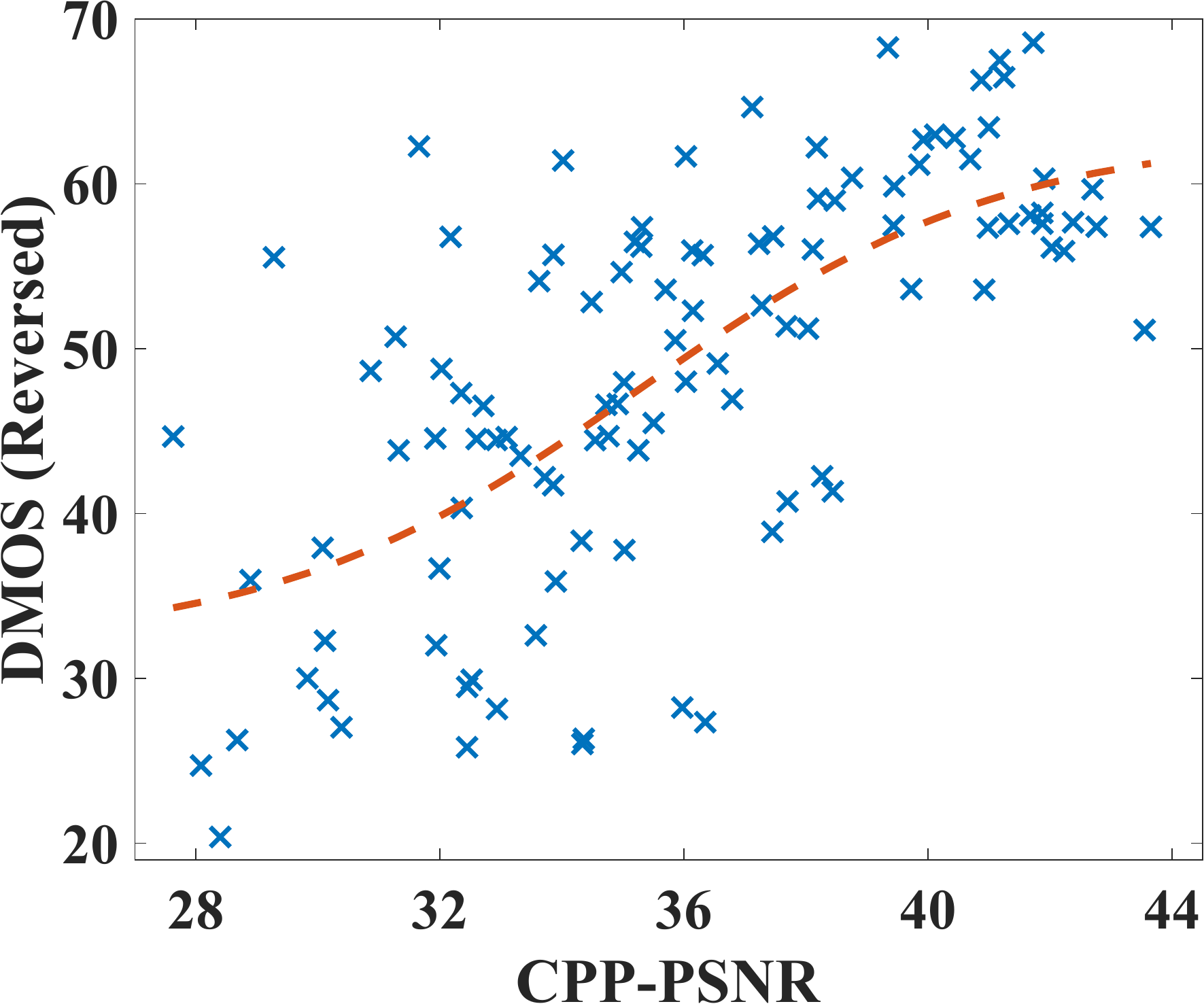}
}
\subfigure[WS-PSNR]{
  \label{fig:scatter:wspsnr}
  \includegraphics[width=0.19\textwidth]{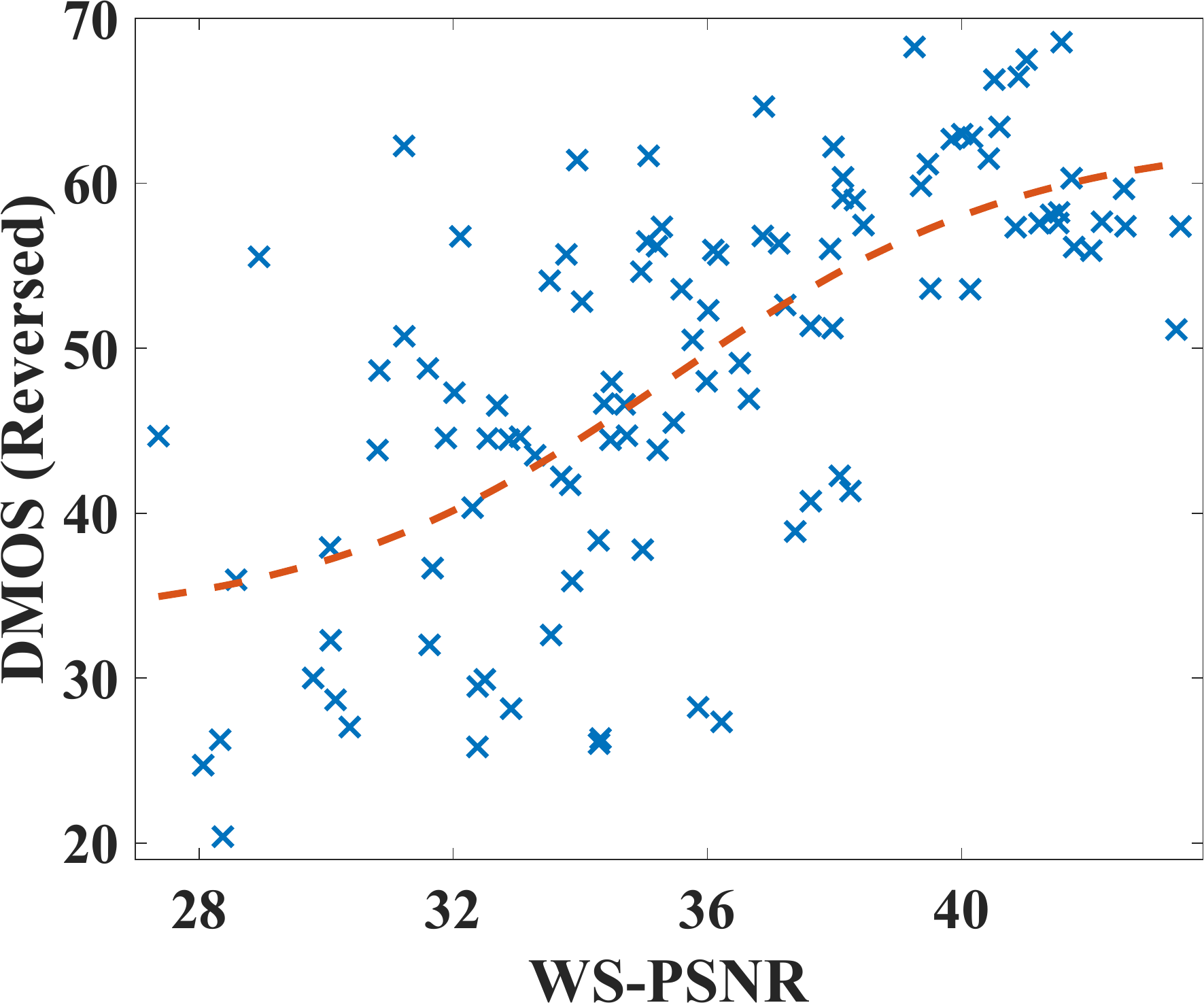}
}
\hspace{-1em}
\end{center}
\vspace{-0.5em}
\caption{Scatter plots for all pairs of objective and subjective VQA scores.}
\label{fig:scatter}
\end{figure*}
\textbf{Loss function.} To train our deep learning model, the loss function $\mathcal{L}$ consists of 3 terms, which is defined by
\begin{equation}
\label{eq:loss}
\footnotesize
\begin{aligned}
\mathcal{L}=&\lambda_1\underbrace{\|\mathbf{s}-\mathbf{s}_\mathrm{g}\|_2^2}_{\text{Mean square error}}\!\!\!+
\lambda_2\underbrace{\frac{1}{nHW}\!\sum_{k=1}^{n}\sum_{(h,v)} \!\!\left(\mathrm{Sobel}_h(\mathbf{M}_k)^2\!+\! \mathrm{Sobel}_v(\mathbf{M}_k)^2\right)^{\frac{3}{2}}}_{\text{Total variation regularization}}
\\&+\lambda_3\underbrace{\|\bm{\beta}\|_2^2.}_{\text{$L_2$ regularization}}
\end{aligned}
\end{equation}
In \eqref{eq:loss}, $\mathbf{s}$ and $\mathbf{s}_\mathrm{g}$ are the vectors of predicted objective VQA scores and the ground truth DMOS scores, for a batch of sequences; $\mathbf{M}_k$ is the $k$-th sensitivity map generated by the CNN component with resolution of $W\times H$; $\text{Sobel}_h$ and $\text{Sobel}_v$ are the Sobel operations alongside horizontal and vertical directions in the pixel coordinate; $\bm{\beta}$ is a vector of all trainable parameters in the deep learning network. In addition, $\lambda_1$, $\lambda_2$ and $\lambda_3$ are the weights of the three terms. In the following, we discuss the three terms of \eqref{eq:loss} in more detail.
\begin{itemize}
  \item Mean square error (MSE). It measures Euclidean distance between the vectors of objective VQA scores and DMOS scores, for a batch of impaired sequences.
  \item Total variation (TV) regularization. It is applied to penalize the high frequency content as an smoothing constraint, since human eyes are insensitive to high frequency details.
  \item The $L_2$ regularization. It is applied to all layers to avoid the overfitting issue in deep learning.
\end{itemize}
\subsection{Evaluation}

\textbf{Settings}. We randomly select 108 impaired sequences of 12 reference sequences as the test set.
The remaining 432 impaired sequences corresponding to 48 reference sequences are as the training set.
In our experiments, all the input sequences are spatially downsampled to the same width of 960 pixels and temporally downsampled with an interval of 45 frames.
This significantly increases the speed of training the deep learning model.
Similar to \cite{kim2017deep}, the learning rate for training our deep learning model is initially set to $5\times10^{-4}$, and the adaptive moment estimation optimizer (ADAM) \cite{kinga2015method} with Nesterov momentum \cite{dozat2016incorporating} is employed.
The deep learning model is trained by 80 epochs, as it is convergent.
The weights of MSE, TV regularization and $L_2$ regularization in the loss function, i.e. $\lambda_1$, $\lambda_2$ and $\lambda_3$ in \eqref{eq:loss}, are set to $1\times10^3$, $1$, and $5\times10^{-3}$, respectively.
Note that all hyperparameters above are tuned over the training set.
In practice, both the HM and EM data are not available for objective VQA on omnidirectional video.
Therefore, the predicted HM maps \cite{xu2017modeling} and EM maps \cite{pan2017salgan} are used in our VQA approach.
Also, we test the performance of our approach with the ground truth HM and EM data, to show the upper bound performance of our approach.

\textbf{Performance evaluation.} Table \ref{tab:result} evaluates the performance of our and other four VQA approaches over the test set.
Among them, DeepQA \cite{kim2017deep} is a state-of-the-art deep learning approach for objective VQA on images, and it is re-trained over our training set of omnidirectional video for fair comparison.
In addition, S-PSNR, CPP-PSNR and WS-PSNR are the latest PSNR related methods for objective VQA on omnidirectional video.
Note that there is no deep learning approach for VQA on omnidirectional video.
For performance evaluation, we measure the correlation between the subjective DMOS scores and objective VQA scores (after non-linear regression) in terms of PCC, SRCC, RMSE and MAE.
We can see from Table \ref{tab:result} that given the predicted HM and EM maps, our VQA approach significantly outperforms other four approaches.
In particular, despite based on the CNN model of \cite{kim2017deep}, our approach is able to increase PCC and SRCC of \cite{kim2017deep} from 0.69 and 0.73 to 0.77 and 0.80, respectively.
Meanwhile, our approach can reduce the RMSE and MAE results of \cite{kim2017deep} by 1.20 and 0.95, respectively.
This is mainly due to the fact that our approach bridges the gap between human behavior and VQA on omnidirectional video, which integrates HM and EM in the deep learning model of VQA.
Additionally, Figure \ref{fig:scatter} shows the scatter plots for all pairs of objective and subjective VQA scores.
In general, intensive scatter points close to the regression curve indicate an effective VQA.
As can be seen in Figure \ref{fig:scatter}, our VQA approach again performs considerably better than other four approaches.

\textbf{Ablation experiments.} Since our VQA approach relies on the HM and EM prediction, we further evaluate the performance of our approach by replacing the predicted HM and EM maps input to the model with ground truth HM and EM maps.
The results are also tabulated in Table \ref{tab:result}.
We can see that when the predicted HM and EM maps are replaced by the ground truth maps, the PCC and SRCC results of our approach can be improved from 0.77 and 0.80 to 0.81 and 0.83, respectively.
This indicates that the accuracy of HM and EM prediction influences the performance of our VQA approach.
On the other hand, this also shows the upper bound performance of our VQA approach, which reaches 0.81, 0.83, 7.96 and 5.86 in terms of PCC, SRCC, RMSE and MAE, respectively.

\begin{table}[!tb]
  \centering
  \caption{Performance comparison between our and other approaches over all 108 sequences of the test set.}
    \resizebox{\linewidth}{!}{
    \begin{tabular}{ccccccc}
    \toprule
    Metrics & \tabincell{c}{Our\\(with ground truth)}  & \tabincell{c}{Our\\(with prediction)} & \tabincell{c}{DeepQA\\ \cite{kim2017deep}} & \tabincell{c}{S-PSNR\\ \cite{yu2015framework}} & \tabincell{c}{CPP-PSNR\\ \cite{zakharchenko2016quality}} & \tabincell{c}{WS-PSNR\\ \cite{sun2017weighted}} \\
    \midrule
    PCC & \textbf{0.81} & 0.78  & 0.69 & 0.69 & 0.68 & 0.67 \\
    SRCC & \textbf{0.83} & 0.80 & 0.73 & 0.70 & 0.69 & 0.68 \\
    RMSE & \textbf{6.91} & 7.38 & 8.53 & 8.54 & 8.67 & 8.77 \\
    MAE & \textbf{5.22} & 5.78 & 6.77 & 6.68 & 6.79 & 6.91 \\
    \bottomrule
    \end{tabular}
    }%
  \label{tab:result}%
\end{table}%
\section{Conclusion and Future Work}
In this paper, we built a large-scale dataset for VQA on omnidirectional video, which has the subjective scores of 600 omnidirectional sequences. Different from other VQA datasets, our dataset also includes the human behavior data of HM and EM on viewing 600 reference/impaired sequences. More importantly, we found from our dataset that subjective quality of omnidirectional video is rather correlated with HM and EM behavior of subjects.
However, the existing approaches have a gap between VQA and human behavior on omnidirectional video. In particular, the state-of-the-art deep learning approaches did not take into account of HM and EM in assessing quality.
Thus, we proposed a deep learning based VQA approaches to seamlessly integrate the HM and EM maps in a deep learning model.
Consequently, the objective scores estimated by our deep learning model were more correlated to the subjective scores than other state-of-the-art approaches, as verified in our experiments.
The promising future work is applying our objective VQA approach in some omnidirectional video processing tasks, for enhancing the perceptual quality of processed video.

\begin{acks}
This work was supported by the National Nature
Science Foundation of China under Grant 61573037 and by the Fok Ying Tung
Education Foundation under Grant 151061.

\end{acks}

\bibliographystyle{ACM-Reference-Format}
\bibliography{sample-bibliography}

\end{document}